%% file: main.tex
\documentclass[10pt,twocolumn,letterpaper]{article}

\usepackage{cvpr}              % To produce the CAMERA-READY version
\input{preamble}

\definecolor{cvprblue}{rgb}{0.21,0.49,0.74}
\usepackage[pagebackref,breaklinks,colorlinks,allcolors=cvprblue]{hyperref}

\usepackage{enumitem}

\def\paperID{13492} % *** Enter the Paper ID here
\def\confName{CVPR}
\def\confYear{2025}

\title{Data Synthesis with Diverse Styles for Face Recognition \\via 3DMM-Guided Diffusion}

\author{
Yuxi Mi$^{1}$\quad
Zhizhou Zhong$^{1}$\quad
Yuge Huang$^{2}$\thanks{Corresponding authors.}\quad
Qiuyang Yuan$^{1}$\quad
Xuan Zhao$^{1}$\quad
Jianqing Xu$^{2}$\\
Shouhong Ding$^{2}$\quad
Shaoming Wang$^{3}$\quad
Rizen Guo$^{3}$\quad
Shuigeng Zhou$^{1}\footnotemark[2]$
\\
$^{1}$ Shanghai Key Lab of Intelligent Information Processing, Fudan University \\
$^{2}$ Youtu Lab, Tencent \quad
$^{3}$ WeChat Pay Lab33, Tencent
\\
{\tt\small \{yxmi20, sgzhou\}@fudan.edu.cn, \{zzzhong22, qyyuan23, xzhao23\}@m.fudan.edu.cn} \\
{\tt\small \{yugehuang, joejqxu, ericshding\}@tencent.com} \\
{\tt\small \{mangosmwang, rizenguo\}@tencent.com}
}

\begin{document}
\maketitle

\input{sec/0_abstract}    
\input{sec/1_intro}
\input{sec/2_related_works}

\input{sec/3_methodology}
\input{sec/4_experiments}

\input{sec/5_conclusion}

% {
%     \small
%     \bibliographystyle{ieeenat_fullname}
%     \bibliography{main}
% }

\input{sec/X_suppl}

{
    \small
    \bibliographystyle{ieeenat_fullname}
    \bibliography{main}
}


\end{document}

%% file: preamble.tex
\usepackage{multirow}

%% file: sec/0_abstract.tex
\begin{abstract}
Identity-preserving face synthesis aims to generate synthetic face images of virtual subjects that can substitute real-world data for training face recognition models. While prior arts strive to create images with consistent identities and diverse styles, they face a trade-off between them. Identifying their limitation of treating style variation as subject-agnostic and observing that real-world persons actually have distinct, subject-specific styles, this paper introduces MorphFace, a diffusion-based face generator. The generator learns fine-grained facial styles, \eg, shape, pose and expression, from the renderings of a 3D morphable model (3DMM). It also learns identities from an off-the-shelf recognition model. To create virtual faces, the generator is conditioned on novel identities of unlabeled synthetic faces, and novel styles that are statistically sampled from a real-world prior distribution. The sampling especially accounts for both intra-subject variation and subject distinctiveness. A context blending strategy is employed to enhance the generator's responsiveness to identity and style conditions. Extensive experiments show that MorphFace outperforms the best prior arts in face recognition efficacy\footnote{Code will be available at https://github.com/Tencent/TFace/.}. 

\vspace{-3mm}

\end{abstract}

%% file: sec/1_intro.tex
\section{Introduction}
\label{sec:intro}

\begin{figure}[tbp]
  \centering
   \includegraphics[width=0.9\linewidth]{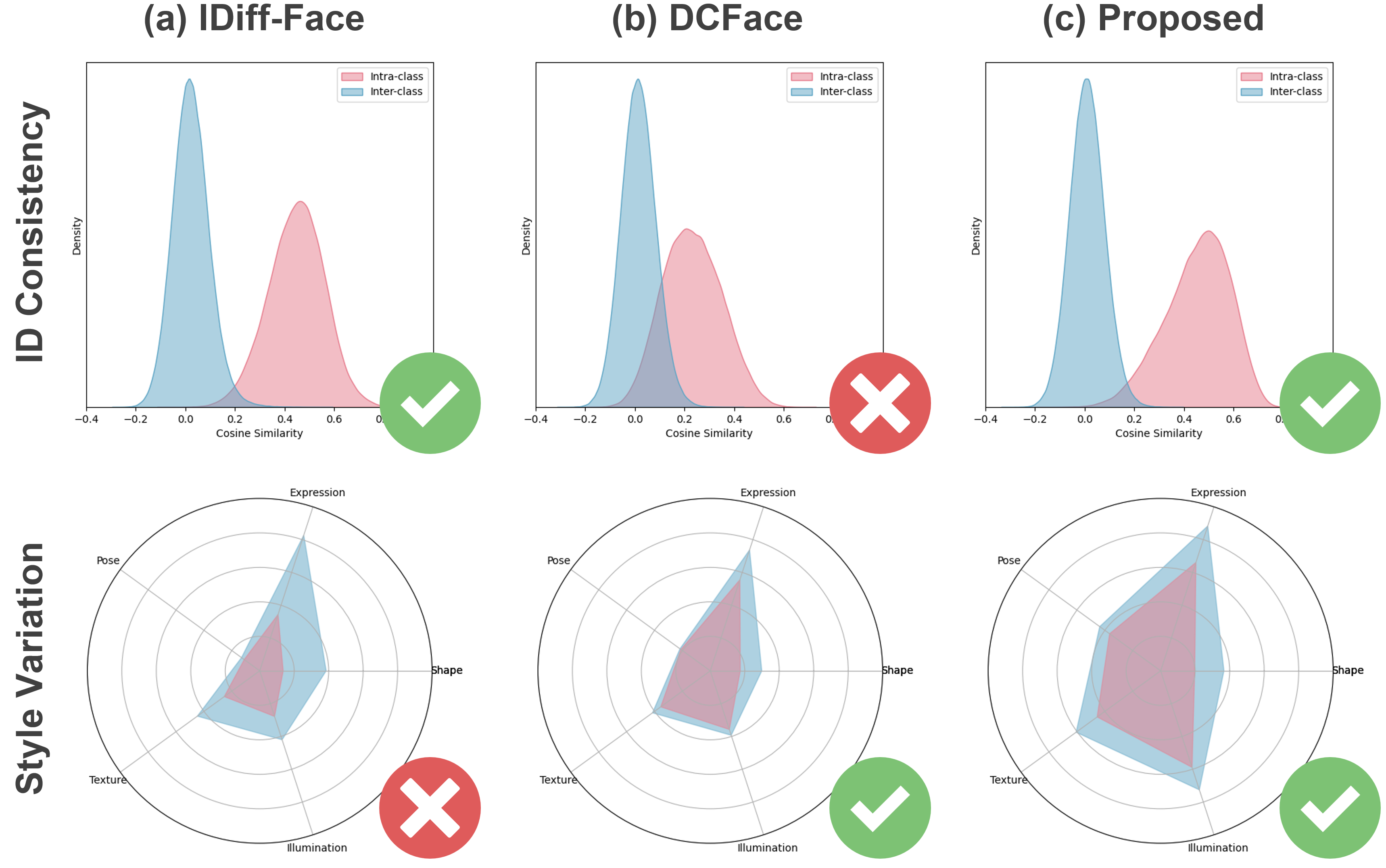}
   \caption{Analyses for identity consistency and style variation across prior arts and our proposed MorphFace. Identity consistency is measured by pairwise cosine similarity and style variation by variances of DECA attributes. Intra-class and inter-class results are represented in red and blue, respectively. Separated curves and a larger shaded area indicate better consistency and variation. Prior arts bear inadequacies in either (a) style variation or (b) identity retention, while (c) MorphFace achieves both goals simultaneously.} 
   \label{fig:paradigm}
   \vspace{-5mm}
\end{figure}

Face recognition (FR) is among the most successful computer vision applications, where persons are identified by model-extracted facial features. FR models are well known for being data-hungry. Their efficacy is built upon large-scale face image training datasets~\cite{cao2018vggface2, zhu2021webface260m, guo2016ms} that contain rich identities and diverse styles, \eg, appearance variations in age, expression and pose. Contemporarily, open-source face image datasets are primarily collected by crawling from the web. The images are potentially enrolled without the informed consent of individuals, which yields serious legal and ethical issues regarding data privacy. 

Identity-preserving face synthesis (IPFS) offers a remedy to the privacy issue. Its objective is to generate face images of virtual subjects and replicate the distribution of real face images so that FR models can be trained on these synthetic faces to effectively recognize real persons. Among previous efforts, early works~\cite{qiu2021synface,boutros2022sface,boutros2024sface2,kolf2023identity,boutros2023exfacegan} are mainly based on generative adversarial networks (GAN) that yet produce face images with limited quality. Recent studies~\cite{boutros2023idiff,kim2023dcface,papantoniou2024arc2face} employ diffusion models (DM) to generate faces of massive unique subjects with fine-grained details. 

The primary challenge of IPFS was to generate multiple faces for the same person. It is recently realized by conditioning a DM's denoising on the person's identity context. We examine the synthetic faces of a related prior art, IDiff-Face~\cite{boutros2023idiff}, in~\cref{fig:paradigm}(a). We measure the cosine similarity between their FR-extracted embeddings and find high \textit{identity consistency} within each subject. Nonetheless, these images are found analogous and lack \textit{style variation} that could help FR generalize. Recent works~\cite{li2024id,kim2023dcface} consider style as an additional DM condition that can be uniformly sampled from external sources, \eg, style banks or pre-trained models. In~\cref{fig:paradigm}(b), we use a DECA~\cite{feng2021learning} 3DMM model to extract style variances of images from DCFace~\cite{kim2023dcface} and observe more varied styles. However, we infer from the similarity metric that their style control negatively impacts identity retention. We refer to this phenomenon as the trade-off between intra-class identity consistency and style variation.

We advocate a paradigm change to create synthetic datasets with both consistent identities and diverse styles. Prior works treat style as a subject-agnostic factor, applying uniform style control across the entire dataset. However, we observe a key divergence from reality in their approach, as they overlook the \textit{distinctiveness of subjects}. In real-world datasets~\cite{cao2018vggface2, zhu2021webface260m, guo2016ms}, images from different subject classes often exhibit distinct styles. For example, individuals from different gender groups typically display different facial shape variations~\cite{li2017learning}. We propose to promote subject distinctiveness in our synthetic faces, which offers two advantages: (1) This enriches dataset variability by combining intra-class style variation with subject-specific styles, without compromising identity consistency; (2) This helps mitigate overfitting to potentially biased styles, allowing FR models to focus on learning identity.

Concretely, we first present a more fine-grained and realistic approach to style control. We use DECA~\cite{feng2021learning} 3DMM to parameterize 3D geometry and facial appearance from an image into attribute sets, and render them into style feature maps. To generate synthetic faces with designated identities and styles, we employ FR-extracted identity embeddings and style feature maps as a DM's context. We employ 3DMM for two reasons: (1) It effectively expresses style in synthetic images; (2) It provides precise, fully parametric control over facial style by adjusting the style attributes. To generate novel faces, we sample style attributes from real-world prior distributions through a \textit{subject-aware strategy}, which explicitly accounts for both intra-class variation and subject distinctiveness. Since we incorporate both identity and style controls during face generation, another key challenge is the effective integration of these two contexts. Based on observations of the DM's denoising process, where styles are primarily established before identity, we propose \textit{context blending} that reweights the style and identity contexts at appropriate denoising timesteps.

We concretize our findings into a novel IPFS generator, MorphFace, named for its ability to morph facial styles through 3DMM renderings. Experimentally, we find that MorphFace achieves a Pareto improvement in balancing intra-class consistency and variation, as shown in~\cref{fig:paradigm}(c). It also significantly enhances FR efficacy, outperforming the best prior methods across all test benchmarks.

This paper presents three-fold contributions:
\begin{itemize}
    \item We present a novel IPFS generator that creates synthetic faces with consistent identities and rich styles. It provides fine-grained style control via 3DMM renderings.
    \item We propose subject-aware sampling that promotes intra-class style variation and subject distinctiveness, and context blending that enhances context expressiveness.
    \item We conduct extensive experiments that demonstrate the state-of-the-art (SOTA) efficacy of our approach.
\end{itemize}

%% file: sec/2_related_works.tex
\section{Related Work}
\label{sec:rw}

\begin{figure*}[tbp]
  \centering
   \includegraphics[width=\linewidth]{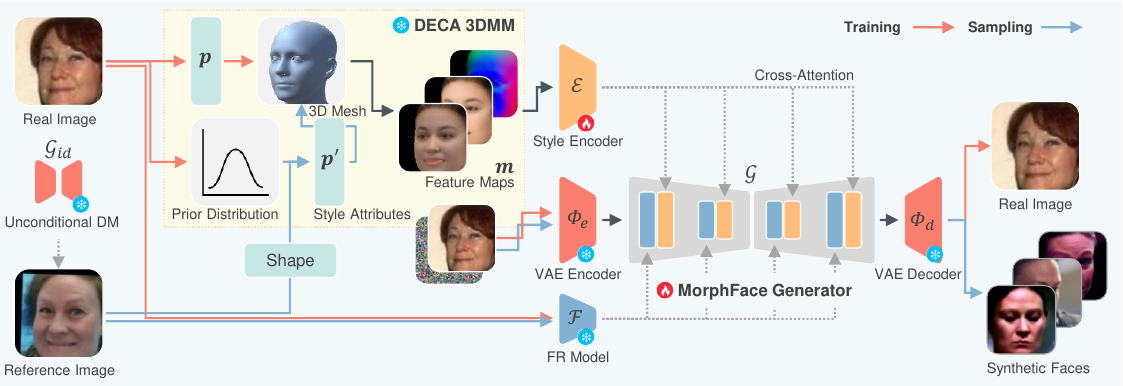}
   \caption{Pipeline of MorphFace. It uses a pair of style and identity contexts to generate faces with designated identity and diverse style. Style is extracted using DECA 3DMM to provides fine-grained, entirely parametric control. To sample virtual faces, unlabeled synthetic images are used as subject reference, and style is sampled statistically for real-world prior distribution. }% The sampling especially accounts for both intra-class style variation and subject distinctiveness.}
   \label{fig:pipeline}
   \vspace{-5mm}
\end{figure*}

\noindent \textbf{Face recognition} aims to match queried face images to an enrolled database. SOTA FR is established on deep neural networks~\cite{he2016deep, boutros2022pocketnet, huang2017densely}, trained using margin-based softmax losses~\cite{boutros2022elastic, deng2019arcface, huang2020curricularface, wang2018cosface, kim2022adaface} on large-scale datasets~\cite{huang2008labeled, cao2018vggface2, zhu2021webface260m, guo2016ms, kemelmacher2016megaface}. Despite the datasets' vital contribution, they often face legal and ethical disputes for being web-crawled without consent~\cite{guo2016ms}. They also exhibit quality problems such as noisy labels and long-tail distributions~\cite{yi2014learning}. FR's performance is measured on benchmark datasets~\cite{zheng2018cross, zheng2017cross, sengupta2016frontal, moschoglou2017agedb} that capture real-world variations, \eg, pose and age.

\noindent \textbf{Face image synthesis} is a long-standing task that has yielded numerous impressive results. Pioneering works use style-based GANs~\cite{karras2021alias,karras2019style,karras2020analyzing,medin2022most,nguyen2019hologan}, 3D priors~\cite{deng2018uv,geng20193d,kim2018deep,medin2022most,nguyen2019hologan,piao2019semi,xu2024chain,huang2023dr}, or semantic attribute annotations~\cite{tran2018representation,shen2018faceid,shen2018facefeat,deng2020disentangled} to generate images with specific facial attributes~\cite{gecer2018semi} or to manipulate existing reference images~\cite{shen2020interpreting}. 
Recent approaches primarily leverage diffusion models~\cite{ho2020denoising,song2020denoising,rombach2022high} to generate subject-conditioned images. Among these, tuning-based methods personalize a pre-trained DM (\eg, Stable Diffusion~\cite{rombach2022high}) on a few images~\cite{ruiz2023dreambooth,gal2022image,ding2023diffusionrig}, extracted features~\cite{hu2021lora,yuan2023inserting,wang2024stableidentity}, or textual descriptions~\cite{gal2023encoder,zhou2023enhancing} of a specific subject, to produce images that reflect that subject's identity. 
Other methods, in contrast, train DMs typically conditioned on subject-descriptive features~\cite{xiao2024fastcomposer,chen2023photoverse,li2024photomaker} from CLIP~\cite{radford2021learning}, FR-extracted identity embeddings~\cite{valevski2023face0,chen2023dreamidentity,peng2024portraitbooth}, or them combined~\cite{yan2023facestudio}. These methods have promoted not only data creation~\cite{chen2024implicit,8972606} but also related tasks~\cite{yan2024df40,yan2024effort,mi2024privacy, zhong2024slerpface}. However, they prioritize high image fidelity over the distinctiveness of subjects. They are less suitable for producing FR training data due to ambiguity in identity retention.

\noindent \textbf{Face recognition with synthetic images} offers benefits in both privacy and quality for FR training~\cite{melzi2024frcsyn,frcsyn2025,deandres2024frcsyn}. Closest to our study, recent works aim to generate multiple synthetic face images for each subject, unseen in real datasets, to replace real images in FR training. We refer to these methods as \textit{identity-preserving face synthesis}. Specifically, SynFace~\cite{qiu2021synface}, SFace~\cite{boutros2022sface}, SFace2~\cite{boutros2024sface2}, IDNet~\cite{kolf2023identity} and ExFaceGAN~\cite{boutros2023exfacegan} use varied GAN architectures~\cite{deng2020disentangled,karras2020training} in subject-conditioned settings, while USynthFace~\cite{boutros2023unsupervised} uses unlabeled images to improve FR training.  DigiFace~\cite{bae2023digiface} utilizes a 3D parametric model to produce distinctive yet less realistic faces.
IDiff-Face~\cite{boutros2023idiff}, DCFace~\cite{kim2023dcface}, Arc2Face~\cite{papantoniou2024arc2face}, CemiFace~\cite{sun2024cemiface} and ID3~\cite{li2024id} are diffusion-based latest works. Most of them~\cite{kim2023dcface,papantoniou2024arc2face,sun2024cemiface,li2024id} explicitly promote style variations during DM's sampling process to improve FR generalization. This paper outperforms them largely by offering more precise and realistic style control.

%% file: sec/3_methodology.tex
\section{Proposed Approach}
\label{sec:method}

% \subsection{Overview}
% \label{subsec:method-overview}

\noindent \textbf{Overview.} We introduce MorphFace, a face generator that produces synthetic face images with consistent identities and varied styles. Our approach is fueled by a latent diffusion model (LDM)~\cite{rombach2022high}. To preserve identity, we condition the LDM on FR-extracted identity embeddings. To vary styles, while prior arts have employed style banks~\cite{kim2023dcface}, similarity metrics~\cite{sun2024cemiface}, and attribute predicates~\cite{li2024id} to coarsely promote style variation, they are unable to control specific style attributes. In contrast, we use 3DMM renderings as the LDM's style contexts. We gain more precise control over style since the renderings provide entirely parametric style descriptions.

To generate unseen face images, we are required to sample novel identities and style contexts. For identity, we obtain reference images of virtual subjects using unlabeled faces from an additional pre-trained DM. For style, we sample 3DMM style attributes in a manner that considers both intra-class style variation and subject distinctiveness, to better mimic real-world style variations. This also differentiates our approach from prior arts~\cite{kim2023dcface,sun2024cemiface,li2024id} which typically apply uniform style control. Experimentally, we find our subject-aware style sampling significantly enhances FR efficacy. We further augment the style and identity contexts during the LDM's certain denoising phases to improve their expressiveness. \Cref{fig:pipeline} illustrates our pipeline. % For style, we sample 3DMM attributes informed by our key observation of real-world distribution: we propose a two-stage strategy—first sampling style representatives for distinct subject classes, then sampling styles for intra-class images based on the representative.

\subsection{Preliminary}
\label{subsec:method-preliminary}

\noindent \textbf{Latent diffusion models}~\cite{rombach2022high} are generative models trained to predict the latent representations $\mathbf{z}$ of input images $\mathbf{x}$ via a gradual denoising process. Let $\mathcal{\phi}_e,\mathcal{\phi}_d$ be a pair of pre-trained encoder and decoder. The image $\mathbf{x}$ is mapped into a latent space as $\mathbf{z}$$=$$\mathcal{\phi}_e(\mathbf{x})$, then is corrupted by variance-controlled Gaussian noise $\mathbf{\epsilon}$ over $0$$\leq$$t$$\leq$$T$ timesteps,  

\begin{equation}
    \label{eq:ldm-noise}
    \mathbf{z}_{t} = \sqrt{\bar{\alpha}_{t}}\mathbf{z}_{0} + \sqrt{1-\bar{\alpha}_{t}}\mathbf{\epsilon},
\end{equation}

\noindent where $\mathbf{z}_{0}$ stands for the clean latent representation, $\alpha_i$ is from a linear variance schedule, and $\bar{\alpha}_t=\prod_{i=1}^{t}{\alpha_i}$. In the denoising process, the model attempts to recover $\mathbf{z}_{t-1}$ iteratively through following transition,

\begin{equation}
    \label{eq:ldm-denoise}
    \mathbf{z}_{t-1} = \frac{1}{\sqrt{\alpha_t}} \left( \mathbf{z}_t - \frac{1 - \alpha_t}{\sqrt{1 - \bar{\alpha}_t}} \mathbf{\epsilon}_{\theta}(\mathbf{z}_t, t, \mathbf{c}) \right) + \sqrt{1-\alpha_t}\mathbf{\epsilon},
\end{equation}

\noindent where $\mathbf{c}$ is a context condition such as identity or style. The image is recovered as $\mathbf{x}$$=$$\mathcal{\phi}_d(\mathbf{z}_0)$. The transition is parameterized by a noise estimator $\mathbf{\epsilon}_{\theta}$ (\eg, U-Net~\cite{ronneberger2015u}) trained with the minimization of an $l_2$ objective,

\begin{equation}
    \label{eq:ldm-train}
    \mathcal{L} = \mathbb{E}_{z_t, t, \mathbf{\epsilon} \sim \mathcal{N}(\mathbf{0}, \mathbf{1})} \left[ \left\| \mathbf{\epsilon} - \mathbf{\epsilon}_\theta(\mathbf{z}_t, t, \mathbf{c}) \right\|_2^2 \right].
\end{equation}

\noindent \textbf{3D morphable face models}~\cite{blanz2023morphable} are parametric models that represent faces in a compact latent space. Among them, FLAME~\cite{li2017learning} uses linear blend skinning to create a 3D mesh of vertices that describes facial geometry, including \textit{shape}, \textit{pose}, and \textit{expression}. DECA~\cite{feng2021learning} incorporates FLAME with additional encoders to further provide facial appearance descriptions, including \textit{texture} and \textit{illumination}, through Lambertian reflectance and spherical harmonics lighting. It produces a set of numerical parameters that determinantly model them as style attributes, which can be rendered into feature maps such as surface normals, albedo, and Lambertian rendering. We use DECA, \textit{wlog.}, as our 3DMM foundation model. For further details, we refer the reader to the latest 3DMM survey paper~\cite{li2024advances}.

\subsection{3DMM-Guided Face Synthesis}
\label{subsec:method-train}

LDM is by design capable of unlabeled face generation. We first condition an LDM $\mathcal{G}$ on identity embeddings to let it generate faces of specific subjects. Concretely, let $\mathbf{X}$ denote the real face image dataset on which we train the LDM. We extract its images' identity embeddings via a pretrained FR model~\cite{boutros2022elastic} $\mathcal{F}$ as $\mathbf{c}_{id}$$=$$\mathcal{F}(\mathbf{x})$, and incorporate $\mathbf{c}_{id}$ into the LDM's training process, \cref{eq:ldm-train}, as context through cross-attention. Notably, this approach is conceptually similar to IDiff-Face~\cite{boutros2023idiff}. \Cref{fig:paradigm}(a) has shown that such generated faces bear insufficiency in intra-class style variation. We consider this as a baseline of the following approach.

We further condition the LDM on 3DMM renderings to promote style variation. 3DMM provides fully parametric descriptions for multiple attributes of facial styles, including shape, expression, pose, texture and illumination. This enables us to precisely control the style of specific face images based on 3DMM's parameters, an unachieved goal of prior arts~\cite{kim2023dcface, sun2024cemiface, li2024id}.

% 方便我们实现具体的控制
% Our approach provides explicit control over facial style. We use 3DMM renderings as an additional context to guide the LDM's generation on specific style attributes. We observe that the 3DMM attributes closely reflect real-world facial styles, certainly including pose, expression, and illumination. Moreover, the temporal changes in a person's appearance, such as changes in body fitness or wrinkles due to aging, can be reflected in shape and texture attributes. We hence can gain fine-grained control over styles via 3DMM. % % including facial shape and expression via principal components, pose (of face and camera) via coordinates, texture via reflectance, and illumination via spherical harmonics lighting. 

\begin{figure}[tbp]
  \centering
   \includegraphics[width=\linewidth]{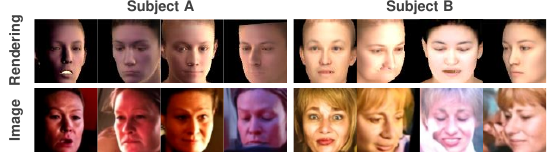}
   \caption{Sample 3DMM feature maps (here, Lambertian renderings) and their synthetic images. \cref{subsec:method-train}: Precise style control and more fine-grained detail can be observed in generated images. \cref{subsec:method-sample}: Sampling subject-aware styles create renderings and images with subjective distinctiveness (\eg, illumination).}
   \label{fig:vis-3dmm}
   \vspace{-5mm}
\end{figure}

Specifically, given input images $\mathbf{x}$, we employ an open-source DECA~\cite{feng2021learning} 3DMM model $\mathcal{M}$ to infer their style attributes, $\mathbf{p}$$=$$\mathcal{M}(\mathbf{x})$. The style attributes are 100,50,9,50,27-dim numerical parameters with human-interpretable meanings for image-wise shape, expression, pose, texture and illumination, respectively. We can concatenate them into a 236-dim vector. Using Lambertian reflectance as part of DECA's integration, we render three feature maps $\mathbf{m}$ entirely parameterized by style attributes $\mathbf{p}$—surface normals, albedo, and Lambertian rendering. The parametric nature will facilitate the sampling of novel styles, illustrated later in~\cref{subsec:method-sample}. From~\cref{fig:pipeline}, we find that the feature maps provide pixel-aligned style descriptions of the input images yet are absence of facial details. We use them to condition the LDM to produce real-looking faces: We concatenate $\mathbf{m}$ along channels and pass them through a simple encoder $\mathcal{E}$ trained end-to-end with the LDM to obtain style embeddings $\mathbf{c}_{sty}$$=$$\mathcal{E}(\mathbf{m})$, and optimize the LDM using both identity and style embeddings as contexts,

\begin{equation}
    \label{eq:ours-train}
    \mathcal{L} = \mathbb{E}_{z_t, t, \mathbf{\epsilon} \sim \mathcal{N}(\mathbf{0}, \mathbf{1})} \left[ \left\| \mathbf{\epsilon} - \mathbf{\epsilon}_\theta(\mathbf{z}_t, t, \mathbf{c}_{id},\mathbf{c}_{sty}) \right\|_2^2 \right].
\end{equation}

To demonstrate our generator's context control,~\cref{fig:vis-3dmm} shows sample synthetic images based on their 3DMM renderings. These images are of high quality and effectively preserve the renderings' style. Unlike prior works, our approach provides explicit, image-wise style control. 

We further distinguish our approach from two close prior arts:  DigiFace~\cite{bae2023digiface} also employs 3DMM for IPFS. However, it directly outputs coarse 3DMM renderings as face images, whereas we incorporate the LDM to generate more realistic faces. DiffusionRig~\cite{ding2023diffusionrig} performs face editing that includes 3DMM as style control. It yet requires burdened subject-wise fine-tuning, and its identity retention is easily nullified upon changing style. It is hence less suitable for IPFS.

\subsection{Synthetic Face Generation}
\label{subsec:method-sample}

\begin{figure}[tbp]
  \centering
   \includegraphics[width=\linewidth]{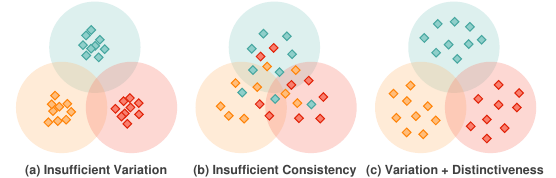}
   \caption{Illustration of style distribution. Regions represent real-world style distributions and diamonds represent samples. (a) Insufficient style variation impairs FR generality. (b) Uniformly sampling styles yields a ``mixed'' distribution that obscure identity consistency. (c) In our proposed approach, style and identity are both promoted by considering the distinctiveness of subjects. } 
   \label{fig:distinctiveness}
   \vspace{-5mm}
\end{figure}

We discuss how to sample novel identities and styles for synthetic face image generation using our trained LDM.

\noindent \textbf{Novel identities.} We employ an unconditional DM $\mathcal{G}_{id}$ to produce unlabeled face images. To improve the images' diversity, we filter them by a cosine similarity threshold of 0.3 on their FR-extracted identity embeddings~\cite{boutros2022elastic} and by image quality assessed via SDD-FIQA~\cite{ou2021sdd}. We use the cleaned images as references for novel subject classes.

\noindent \textbf{Novel styles.} Since the feature maps $\mathbf{m}$ are entirely parameterized by style attributes $\mathbf{p}$, we can produce novel styles by sampling new style attributes $\mathbf{p}'$. To mimic real-world style variations, we propose to sample $\mathbf{p}'$ statistically from the prior style distribution of LDM training dataset. Formally, let $\mathbf{P}$$=$$\mathcal{M}(\mathbf{X})$ be the style attribute set of $\mathbf{X}$, and $\mathbb{D}(\mathbf{P})$ be its distribution. The general form of sampling $\mathbf{p}'$ is as 

\begin{equation}
    \label{eq:sample-general}
    \mathbf{p}'\in\mathbf{P}', \quad \mathbf{P}'\sim \mathbb{D}(\mathbf{P}).
\end{equation}

\noindent We note that $\mathbb{D}(\mathbf{P})$ can be approximated as a multiplicative Gaussian distribution, \ie, $\mathbb{D}(\mathbf{P})$$\sim$$\mathcal{N}(\mathbf{\mu},\mathbf{\Sigma})$, where $\mathbf{\mu}$ and $\mathbf{\Sigma}$ represent the mean and covariance matrix of $\mathbf{P}$. This approximation is grounded by the nature of 3DMM~\cite{blanz2023morphable} and prior studies' findings~\cite{paysan20093d,booth2018large}, and is empirically validated. We leave further discussion to the supplementary material.

\Cref{eq:sample-general} does not specify how each $\mathbf{p}'$ is sampled from $\mathbf{P}'$. Prior arts~\cite{kim2023dcface,li2024id,sun2024cemiface} mainly offer uniform sampling, \ie, providing subject-agnostic style context to each synthetic image. Similarly, we can uniformly sample styles by rewriting~\cref{eq:sample-general} as $\mathbf{p}'$$\sim$$\mathcal{N}(\mathbf{\mu}, \mathbf{\Sigma})$. However, in~\cref{subsec:exp-sample}, we find this means yields suboptimal FR efficacy. %, and the synthetic images preserve less consistent identities.

We propose an improved strategy to better replicate real-world style variations by considering both \textit{intra-class style variation} and \textit{style distinctiveness of subjects}. Intra-class style variation imposes a seeming dilemma: Its insufficiency may impair FR generality~\cite{boutros2023idiff}, yet its excessiveness also reduces FR efficacy since this may obscure the retention of identities~\cite{kim2023dcface}, as illustrated in~\cref{fig:distinctiveness}.

While prior works advocate uniform style variations, our key observation from real-world datasets~\cite{yi2014learning,guo2016ms,zhu2021webface260m} reveals that each subject actually exhibits style distinctiveness that should be considered. For instance, women and men often possess different facial shapes~\cite{li2017learning}; A juvenile may have more youthful photos enrolled in a dataset than an elderly individual, creating age-related distinctions. We believe that such subject-specific distinctiveness plays a crucial role in dataset quality: It enhances dataset variability with less negative impacts on identity consistency, and helps FR models mitigate potential overfitting on biased styles.

We propose \textit{subject-aware style sampling}, concretized from~\cref{eq:sample-general}, based on the observation. To address subject distinctiveness, we first sample class-wise distribution from the style attribute set $\mathbf{P}'$. Then, we sample image style from its class distribution to allow intra-class variation. Formally, let $\mathbf{P}'$$=$$\bigcup_{i=1}^{m} \mathbf{P}'_i$ be a division of $\mathbf{P}'$, where $m$ is the number of unique subjects. We sample $\{\mathbf{P}'_i\}_{i\in[m]}$ as

\begin{equation}
    \mathbf{P}'_i \sim \mathcal{N}(\mathbf{\mu}_i, \mathbf{\Sigma}_i), \quad \sum_{i=1}^{m} \gamma_i \mathcal{N}(\mathbf{\mu}_i, \mathbf{\Sigma}_i) = \mathcal{N}(\mathbf{\mu}, \mathbf{\Sigma}),
\end{equation}

\noindent where $\sum_{i}\gamma_i$$=$$1$. Each class's $\mathbf{\mu}_i$ and $\mathbf{\Sigma}_i$ vary by real-world distributions of class means and covariances. We then sample $\mathbf{p}'$ from class-wise distribution,
% if each class contains an equal number of images

\begin{equation}
    \mathbf{p}' \in \mathbf{P}'_i, \quad \mathbf{p}' \sim \mathcal{N}(\mathbf{\mu}_i, \mathbf{\Sigma}_i).
\end{equation}

% Currently, all style attributes including shape are sampled at random. 
Additionally, we find that using facial geometry similar to the subject's reference image can improve identity consistency. It is achieved by replacing the intra-class mean $\mathbf{\mu}_i$ of facial shape attributes with the reference image's ground truth. In~\cref{fig:vis-3dmm}, we produce feature maps that vary adequately within each subject and more significantly across subjects, better reflecting real-world scenarios.

\subsection{Context Blending}
\label{subsec:method-blend}

\begin{figure}[tbp]
  \centering
   \includegraphics[width=\linewidth]{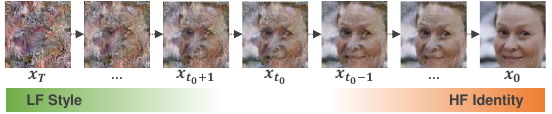}
   \caption{During denoising, LF styles (\eg, pose and shape) are earlier established than HF identity details by the nature of DM.  We augment style and identity contexts before and after a shifting timestep $t_0$ via CFG, respectively, to improve their expressiveness.} 
   \label{fig:denoising}
   \vspace{-5mm}
\end{figure}

As $\mathcal{G}$ is conditioned on both identity and style contexts, we discuss their effective integration. We empirically find that the guidance of $\mathbf{c}_{id}$ and $\mathbf{c}_{sty}$ can slightly contradict each other during image generation, due to the inherent tension between identity and style. To demonstrate, later in~\cref{subsec:exp-blend}, we separately strengthen $\mathbf{c}_{id}$ or $\mathbf{c}_{sty}$ via classifier-free guidance~\cite{ho2022classifier} (CFG), an inference-time method for context augmentation, and find the generated images exhibit reduced style variation and identity consistency, respectively.

% We further find that the identity and style contexts, $\mathbf{c}_{id}$ and $\mathbf{c}_{sty}$, can potentially contradict each other in image generation due to the inherent tension between identity consistency and style variation. To analyze, we adjust their influence on generation by applying 0–50\% dropout to the feature dimensions of either $\mathbf{c}_{id}$ or $\mathbf{c}_{sty}$. We examine the generated images by the cosine similarity of identity embeddings and by the average variance of style attributes, extracted via DECA. From {\color{blue}Fig. X}, we observe a slight trade-off.
To improve the contexts' expressiveness, we investigate DM's denoising process from a frequency perspective. DMs are known to favor specific frequency components at certain denoising timesteps: Low-frequency (LF) components are emphasized in early timesteps, while high-frequency (HF) details are progressively refined~\cite{qian2024boosting, ruhe2024rolling}. In our generator, identity and style contexts align with HF and LF features, respectively: Prior works indicate that facial identity $\mathbf{c}_{id}$ is largely captured with HF details~\cite{wang2022privacy,mi2022duetface}, while $\mathbf{c}_{sty}$ mainly consists coarse LF features from 3DMM rendering. As shown in~\cref{fig:denoising}, step-wise denoising reveals that styles (\eg, pose and illumination) are established very early, while facial identity emerges in later steps.

% We study the LDM's denoising of faces from a frequency perspective. DMs are known for their preference of certain frequency components at certain denoising timesteps: They always prioritize the images' low-frequency (LF) components in early timesteps, then progressively move to high-frequency (HF) details~\cite{qian2024boosting,ruhe2024rolling}.  Mean while, our identity and style contexts can be considered  as representative of HF and LF features: Prior arts find facial identity  $\mathbf{c}_{id}$ is mainly represented HF details ~\cite{wang2022privacy,mi2022duetface}, and $\mathbf{c}_{sty}$ is extracted from 3DMM rendering that consists coarse LF facial features. Combining them, in {\color{blue}Fig. X}, we visualize the step-wise denoising process and find styles (\eg, pose and illumination) are largely primarily established before identities.

\begin{small}
\begin{table*}[tbp]
\centering
\begin{tabular}{lllcccccc}
\toprule
\textbf{Method}    & \textbf{Venue} & \textbf{Volume (IDs × imgs)} & \textbf{LFW}   & \textbf{CFP-FP} & \textbf{AgeDB} & \textbf{CPLFW} & \textbf{CALFW} & \textbf{Avg.} \\
\midrule
CASIA              & (real)         & 0.49M (10.5K × 47)           & 99.38          & 96.91           & 94.50          & 89.78          & 93.35          & 94.79            \\
\midrule
SynFace            & ICCV 21      & 0.5M (10K × 50)              & 91.93          & 75.03           & 61.63          & 70.43          & 74.73          & 74.75            \\
SFace              & IJCB 22      & 0.6M (10K × 60)              & 91.87          & 73.86           & 71.68          & 77.93          & 73.20          & 77.71            \\
DigiFace           & WACV 23      & 0.5M (10K × 50)              & 95.40          & 87.40           & 76.97          & 78.87          & 78.62          & 83.45            \\
IDnet              & CVPR 23      & 0.5M (10K × 50)              & 84.83          & 70.43           & 63.58          & 67.35          & 71.50          & 71.54            \\
DCFace             & CVPR 23      & 0.5M (10K × 50)              & 98.55          & 85.33           & 89.70          & 82.62          & 91.60          & 89.56            \\
IDiff-Face         & ICCV 23      & 0.5M (10K × 50)              & 98.00          & 85.47           & 86.43          & 80.45          & 90.65          & 88.20            \\
ExFaceGAN          & IJCB 23      & 0.5M (10K × 50)              & 93.50          & 73.84           & 78.92          & 71.60          & 82.98          & 80.17            \\
SFace2             & BIOM 24      & 0.6M (10K × 60)              & 94.62          & 76.24           & 74.37          & 81.57          & 72.18          & 79.80            \\
Arc2Face           & ECCV 24      & 0.5M (10K × 50)              & 98.81          & 91.87           & 90.18          & 85.16          & 92.63          & 91.73            \\
ID3                & NeurIPS 24   & 0.5M (10K × 50)              & 97.68          & 86.84           & 91.00          & 82.77          & 90.73          & 89.80            \\
CemiFace           & NeurIPS 24   & 0.5M (10K × 50)              & 99.03          & 91.06           & 91.33          & 87.65          & 92.42          & 92.30            \\
\textbf{MorphFace} & (ours)         & 0.5M (10K × 50)              & \textbf{99.25} & \textbf{94.11}  & \textbf{91.80} & \textbf{88.73} & \textbf{92.73} & \textbf{93.32}   \\
\midrule
DigiFace           & WACV 23      & 1.2M (10K × 72, 100K × 5)   & 96.17          & 89.81           & 81.10          & 82.23          & 82.55          & 86.37            \\
DCFace             & CVPR 23      & 1.2M (20K × 50, 40K × 5)    & 98.58          & 88.61           & 90.07          & 85.07          & 92.82          & 91.21            \\
Arc2Face           & ECCV 24      & 1.2M (20K × 50, 40K × 5)    & 98.92          & 94.58           & 92.45          & 86.45          & 93.33          & 93.15            \\
\textbf{MorphFace} & (ours)         & 1.2M (24K × 50)              & \textbf{99.35} & \textbf{94.77}  & \textbf{93.27} & \textbf{90.07} & \textbf{93.40} & \textbf{94.17}  \\
\bottomrule
\end{tabular}
\caption{Comparsion with SOTAs by FR recognition accuracy. Our proposed MorphFace outperforms SOTAs on all benchmarks.}
\label{tab:exp-sota}
\end{table*}
\end{small}

Based on the observation, we propose \textit{context blending} to enhance the guidance of either context at its appropriate denoising timesteps. Specifically, we strengthen $\mathbf{c}_{sty}$ in earlier timesteps and $\mathbf{c}_{id}$ in later timesteps to improve the LDM's responsiveness to these contexts. Formally, during training, we first probabilistically replace $\mathbf{c}_{id}$ and $\mathbf{c}_{sty}$ with learnable empty contexts $\mathbf{c}_{id}^{\emptyset}$ and $\mathbf{c}_{sty}^{\emptyset}$; during inference time, we employ CFG for context augmentation. We rewrite~\cref{eq:ldm-denoise} in CFG-form as

\begin{equation}
    \label{eq:ldm-denoise-rewrite}
    \mathbf{z}_{t-1} = \frac{1}{\sqrt{\alpha_t}} \left( \mathbf{z}_t - \frac{1 - \alpha_t}{\sqrt{1 - \bar{\alpha}_t}} \mathbf{\epsilon}_{cfg} \right) + \sqrt{1-\alpha_t}\mathbf{\epsilon},
\end{equation}

\noindent where $\mathbf{\epsilon}_{cfg}$ is weighted by $w$ as

\begin{equation}
    \mathbf{\epsilon}_{cfg}=(1+w)\mathbf{\epsilon}_{\theta}(\mathbf{z}_t, t, \mathbf{c}_{id},\mathbf{c}_{sty}) - w\mathbf{\epsilon}_{t}.
\end{equation}

\noindent We choose a time-varying $\mathbf{\epsilon}_{t}$ as $\mathbf{\epsilon}_{\theta}(\mathbf{z}_t, t, \mathbf{c}_{id},\mathbf{c}_{sty}^{\emptyset})$ for $t\in(t_0,T]$ to augment style, and as $\mathbf{\epsilon}_{\theta}(\mathbf{z}_t, t, \mathbf{c}_{id}^{\emptyset},\mathbf{c}_{sty})$ for $t\in[0,t_0]$ to augment identity, where $t_0$ is a ``shifting'' timestep. \Cref{subsec:exp-blend} shows that context blending improves identity consistency and style variation, and enhances FR efficacy.

%  % {\color{blue} Figure X(x)} compares images generated with and without context blending. We observe better quality and more facial details (\eg, wrinkles) in blended images.
% (\mathbf{z}_t, t, \mathbf{c}_{cfg})
% % 就写一行，下面解释在不同的timestep以空置换条件
% \begin{small}
% \begin{equation}
%     \begin{cases}
%         (1+w)\mathbf{\epsilon}_{\theta}(\mathbf{z}_t, t, \mathbf{c}_{id},\mathbf{c}_{sty}) - w\mathbf{\epsilon}_{\theta}(\mathbf{z}_t, t, \mathbf{c}_{id}),~t\in(t_0,T],\\
%         (1+w)\mathbf{\epsilon}_{\theta}(\mathbf{z}_t, t, \mathbf{c}_{id},\mathbf{c}_{sty}) - w\mathbf{\epsilon}_{\theta}(\mathbf{z}_t, t, \mathbf{c}_{sty}),~ t\in[0,t_0].
%     \end{cases}
% \end{equation}    
% \end{small}

% To reconcile the trade-off, we first observe DM's denoising from a frequency perspective. Latest studies~\cite{qian2024boosting,ruhe2024rolling} reveal that DM always prioritizes the recovery of images' low-frequency (LF) components in early timesteps, then progressively moves to high-frequency (HF) details. Coincidentally, we find $\mathbf{c}_{sty}$ and $\mathbf{c}_{id}$ to contain mainly 

% 我们从denoising的原理着手，更好地增强二者的expressiveness。我们注意到style本身是coarse的feature map，包含更多低频信息，而identity则包含更多高频信息。Diffusion的原理决定了在去噪时，高频信息首先被破坏，而它将先重建低频信息，再重建高频信息。图X展示了不同去噪时间步的示例图像，我们注意到，图像的整体style如pose远先于identity构建。

%% file: sec/4_experiments.tex
\section{Experiments}
\label{sec:exp}

\subsection{Experimental Setup}
\label{subsec:exp-setup}

\noindent \textbf{Datasets.} We train our LDM $\mathcal{G}$ on CASIA-WebFace~\cite{yi2014learning}, a dataset that consists of 490k quality-varying face images from 10575 identities. We benchmark our FR model $\mathcal{F}_{syn}$ on 5 widely used test datasets, LFW~\cite{lfwtechupdate}, CFP-FP~\cite{sengupta2016frontal}, AgeDB~\cite{moschoglou2017agedb}, CPLFW~\cite{zheng2018cross}, and CALFW~\cite{zheng2017cross}. CFP-FP and CPLFW are designed to measure the FR in cross-pose variations, and AgeDB and CALFW are for cross-age variations.

% We postpone implementation details to the supplementary material due to space limitation.
% \noindent \textbf{Implementation details.} We use publicly released FR model $\mathcal{F}$ from~\cite{kim2023dcface}, unconditional DM $\mathcal{G}_{id}$ from~\cite{ho2020denoising}, and encoder and decoder $\mathcal{\phi}_e,\mathcal{\phi}_d$ from ~\cite{rombach2022high}. For the style encoder $\mathcal{E}$, we employ 2 convolutional layers plus a linear layer. We train our generator $\mathcal{G}$ for 250K steps, using an Adam optimizer~\cite{kingma2014adam}, an initial learning rate of 1e-4, and a total batch size of 512. To evaluate our synthetic dataset, we train an IR-50~\cite{duta2021improved} FR model $\mathcal{F}_{syn}$ for 40 epochs using an SGD optimizer~\cite{ruder2016overview}, a total batch size of 256, and an initial learning rate of 0.1. We run all experiments on 8 NVIDIA RTX 3090 GPUs. We choose $t_0$=500 and $w$=0.5. 

% \noindent \textbf{Critical feature shapes.} $\mathcal{G},\mathcal{G}_{id}$ produces 3×128×128 images, and so is each of the 3DMM feature maps. The latent representation of $\mathcal{G}$ is 3×32×32. FR inputs are resized 3×112×112 images. The lengths of $\mathbf{c}_{id}$ and $\mathbf{c}_{sty}$ are 512.

\subsection{Comparison with SOTAs}
\label{subsec:exp-sota}

\begin{figure}[tbp]
  \centering
   \includegraphics[width=\linewidth]{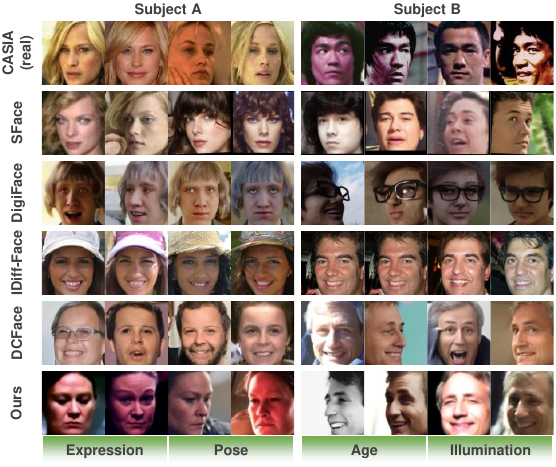}
   \caption{Image visualization for MorphFace and SOTAs. Our approach produces faces with intra-class variation and subject distinctiveness of style. It better replicates real-world style variations. } 
   \label{fig:vis-sota}
   \vspace{-5mm}
\end{figure}

% We compare MorphFace with IPFS SOTAs.  我们synthesize了两个不同volume的数据集1. 常见的0.5M，2. 更大的1.2M，我们对每个身份sample 50张图片。 In Tab.1, we report the test accuracy of our FR model $\mathcal{F}_{syn}$ and SOTAs on 5 widely used test datasets. We have discussed these SOTAs in~\cref{sec:rw}. We also report the result on CASIA-WebFace, \ie, our LDM training data. Note that some SOTAs may have larger training datasets for the generator (Arc2Face), larger FR backbone (CemiFace), and real-world reference images (DCFace) that could benefit their results. We highlight several key points: (1) MorphFace clearly outperforms all SOTAs on all benchmarks for both data volumes; Notably, we outperform the best SOTA for 2.24 on CFP-FP, 1.08 on CPLFW, and 1.02 on average. As CFP-FP and CPLFW are both pose-varying datasets, we believe our proposed method is especially effective to improve pose variations. (2) Our 0.5M setting outperforms the 1.2M setting of SOTAs, reflecting the high-quality and strong generality of our propose approach. (3) Our 1.2M setting achieved on-par performance on CPLFW and CALFW with the real-world CASIA baseline. (4) Diffusion-based methods 都取得了不错的结果，这说明diffusion在生成更具有可保持身份细节的图像上可能具有优势。

\noindent \textbf{Recognition accuracy.} We generate synthetic datasets using trained $\mathcal{G}$. We synthesize 2 data volumes: 0.5M/1.2M face images from 10K/24K subjects with 50 images for each subject. We train an IR-50 FR model $\mathcal{F}_{syn}$ on our synthetic datasets and compare IPFS SOTAs~\cite{qiu2021synface,boutros2022sface,boutros2024sface2,kolf2023identity,boutros2023exfacegan,boutros2023idiff,kim2023dcface,papantoniou2024arc2face,sun2024cemiface,li2024id} discussed in~\cref{sec:rw}. We benchmark them on 5 widely used test datasets by FR recognition accuracy in~\cref{tab:exp-sota}. Note that some SOTAs may have larger datasets for the generator~\cite{papantoniou2024arc2face}, larger FR backbones~\cite{sun2024cemiface}, and real-world reference images~\cite{kim2023dcface} that could benefit their results. % We also report the result on CASIA-WebFace~\cite{yi2014learning} on our LDM training data

We highlight several key points: (1) MorphFace outperforms \textit{all} SOTAs on \textit{all} test datasets for both 0.5M/1.2M volumes. Notably, we outperform the best SOTA for 2.24 on CFP-FP, 1.08 on CPLFW, and 1.02 on average. As CFP-FP and CPLFW are both pose-varying datasets, this suggests MorphFace could be especially beneficial for cross-pose settings. (2) Our average result of 0.5M outperforms the 1.2M results of SOTAs, demonstrating our approach's high capability. (3) Our 1.2M result achieves on-par performance on CPLFW and CALFW with the real-world CASIA~\cite{yi2014learning}. (4) DM-based methods~\cite{papantoniou2024arc2face,sun2024cemiface,kim2023dcface,boutros2023idiff,li2024id} all exhibit quite satisfactory FR efficacy, which may be attributed to better generations of identity-reflecting HF facial details.

\noindent \textbf{Visualization.} We compare CASIA, several SOTAs that released their datasets, and MorphFace. In~\cref{fig:vis-sota}, we sample 8 images of 2 subjects from each dataset. We highlight: (1) \textbf{SFace}~\cite{boutros2022sface} preserves less consistent identities; (2) \textbf{DigiFace}~\cite{bae2023digiface} directly uses 3DMM renderings as images, producing less realistic faces. It yet better represents accessories (\eg, glasses); (3) \textbf{IDiff-Face}~\cite{boutros2023idiff} lacks intra-class variation, producing mainly frontal faces; (4) \textbf{DCFace}~\cite{kim2023dcface} largely promotes style variation. However, some attributes (\eg, expression) are replicated across its subjects, suggesting less distinctiveness and overfitting to biased styles. It also occasionally creates artifacts (\eg, gender transition) due to degraded identity consistency; (5) \textbf{MorphFace} promotes intra-class style variations including expression, pose, age and illumination, and also creates more distinctive subjects. It better mimics the style variations of real-world datasets.

% For those SOTAs that have open-source synthetic images, we visualize them. 我们也visualize了CASIA。 Notably, SFace怎么样；DigiFace generates less realistic Faces；特别地，我们注意到IDiff-Face的图像缺乏类内变化；DCFace的图像（我们可视化了它的两个subjects）尽管具有丰富的类内变化，但类间区分性却不明显，表现为近似的色调和表情；我们的方法既具有丰富的类内变化（pose、表情、光照）、也具有类的独特性。这更加接近真实数据集，解释了我们方法的良好表现。

\begin{figure}[tbp]
  \centering
   \includegraphics[width=0.9\linewidth]{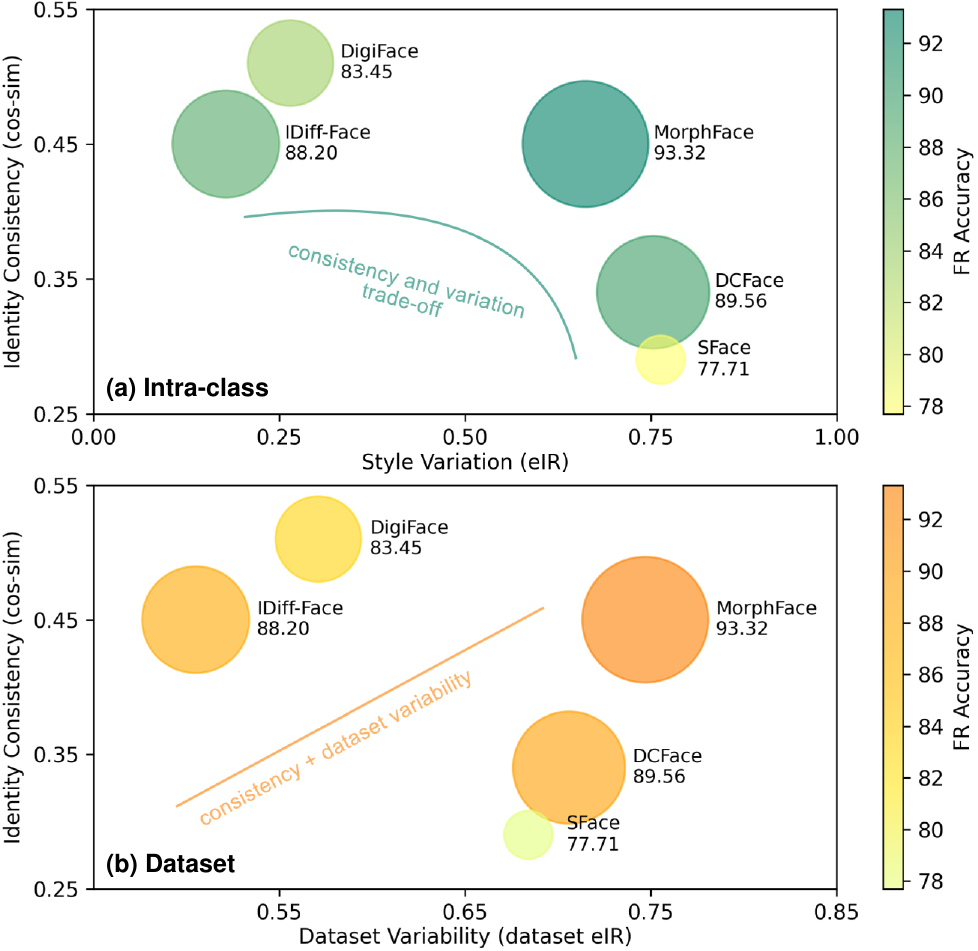}
   \caption{Comparison among 5 methods by consistency and variation metrics. Circle colors and sizes depict FR accuracy. (a) MorphFace outperforms SOTAs in consistency and variation trade-off. (b) It promotes both consistency and datasets' overall variability.} 
   \label{fig:trade-off}
   \vspace{-4mm}
\end{figure}

\noindent \textbf{Consistency \textit{vs.} Variation.} We quantitatively investigate the balance between intra-class identity consistency and style variation. We calculate the extended Improved Recall~\cite{kynkaanniemi2019improved} (eIR) metric from~\cite{kim2023dcface} on intra-class images to measure style variation. It captures the sparseness of style space manifolds where larger eIR stands for more diverse styles. We measure identity consistency by the average cosine similarity between identity embedding pairs. In~\cref{fig:trade-off}(a), we compare the similarity and eIR among MorphFace and SOTAs~\cite{boutros2022sface, bae2023digiface, boutros2023idiff, kim2023dcface}, where the FR accuracy is depicted by the color and size of circles. We observe a clear trade-off between consistency and variation. While SOTAs either prioritize identity or style, our approach seeks a balance that improves FR efficacy.

\noindent \textbf{Consistency \& Dataset variability.} Dataset's overall variability is a combined effort of intra-class variation and subject distinctiveness. By promoting distinctiveness, we can improve variability with less impact on consistency. In~\cref{fig:trade-off}(b), we measure variability by dataset-wise eIR. MorphFace manages to create both consistent identities and diverse styles from a dataset perspective. This explains its better performance as both factors are vital for FR efficacy.

% 我们已经在前文详细地讨论了该trade-off（注意到该trade-off是类内的），简单地说，对于既往方法，强调身份一致性易导致类内style单一，而uniform style sampling容易导致类内的身份信息被过于多样的style破坏（如呈现性别变化）。这反映为trade-off。具体而言，我们使用预训练的FR模型提取身份模板并计算相似度，以类内平均相似度作为consistency的指标；我们沿用DCFace的方法，以Improved Recall作为类内多样性指标。图上，我们以圆的大小和颜色标记recognition accuracy。可以看到我们的方法取得了最好的trade-off，和最高的accuracy。

% 我们缺少一个数据集整体多样性的指标，要补吗？

\subsection{Effect of Style Sampling Strategy}
\label{subsec:exp-sample}

\begin{figure}[tbp]
  \centering
   \includegraphics[width=\linewidth]{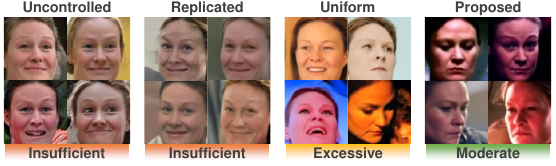}
   \caption{Sample synthetic faces from 4 different style sampling strategies. While others provide insufficiently or excessively varied styles, our proposed approach offers moderate style variation.} 
   \label{fig:style-strategy}
   \vspace{-3mm}
\end{figure}

\begin{small}
    \begin{table}[tbp]
    \centering
\begin{tabular}{llccc}
\toprule
& \textbf{Strategy}      & \textbf{eIR} & \textbf{cos-sim} & \textbf{FR Avg.} \\

\midrule
\multirow{4}{*}
{\begin{tabular}[c]{@{}l@{}}\textbf{(a)}\\ \textbf{Style} \end{tabular}} 
& Uncontrolled           & 0.475        & 0.41             & 91.59            \\
& Replicated             & 0.178        & \textbf{0.61}             & 82.60            \\
& Uniform                & \textbf{0.720}        & 0.33             & 92.41            \\
& \textbf{Proposed} & 0.642        & 0.45             & \textbf{93.32}            \\
\midrule
\multirow{4}{*}{\begin{tabular}[c]{@{}l@{}}\textbf{(b)} \\ \textbf{Context}\end{tabular}}
& W/o blending           & 0.608        & 0.37             & 93.11            \\
& W/ identity           & 0.575        & \textbf{0.51}             & 92.75            \\
& W/ style           & \textbf{0.687}       & 0.35             & 92.83            \\
& \textbf{W/ blending}   & 0.642        & 0.45             & \textbf{93.32}            \\
\bottomrule
\end{tabular}
    \caption{Analyses of identity consistency, style variation and FR efficacy for style sampling and context blending strategies.}
    \label{tab:exp-sample}
    \end{table}
    \vspace{-2mm}
\end{small}

\textit{How does the style sampling strategy affect identity consistency, style variation, and FR efficacy?} We compare 4 settings: (1) \textbf{Uncontrolled}, which we condition the generator $\mathcal{G}$ solely on $\mathbf{c}_{id}$, similar to~\cite{boutros2023idiff}; (2) \textbf{Replicated}, which we reuse the style feature maps of the reference image, instead of sampling novel styles; (3) \textbf{Uniform}, which we sample styles uniformly like~\cite{kim2023dcface} as $\mathbf{p}'$$\sim$$\mathcal{N}(\mathbf{\mu}, \mathbf{\Sigma})$; (4) \textbf{Proposed}, our subject-aware style sampling discussed in~\cref{subsec:method-sample}.

\noindent \textbf{Visualization.} \Cref{fig:style-strategy} shows sample synthetic images based on the same reference image from 4 settings. We observe: (1) Uncontrolling yields insufficient style variation; (2) Replicating the reference image's style results in even less variation as the style is negatively controlled by the same $\mathbf{c}_{sty}$; (3) Though uniform sampling promotes more diverse styles, its variation is sometimes excessive for the same subject and could affect identity retention; (4) Our subject-aware setting offers moderate style variation.

\noindent \textbf{Quantitative analysis.} In~\cref{tab:exp-sample}(a), we present results on eIR, cosine similarity, and average FR accuracy. The low eIR of uncontrolled settings suggests insufficient style variation. We observe a significant trade-off between replicated and uniform settings, which both yield suboptimal performance. The subject-aware setting offers the best FR efficacy due to balanced consistency and variation.

% 想办法和subjectiveness关联一下

% 比较4个方法，ours，仅身份的baseline、逆向控制身份的（模拟IDiff-Face的worse情况）、均一采样的，报告其在5个数据集上的平均accuracy、IR和余弦相似度。有必要的话拉一个可视化的图

\subsection{Effect of Context Blending}
\label{subsec:exp-blend}

\begin{figure}[tbp]
  \centering
   \includegraphics[width=\linewidth]{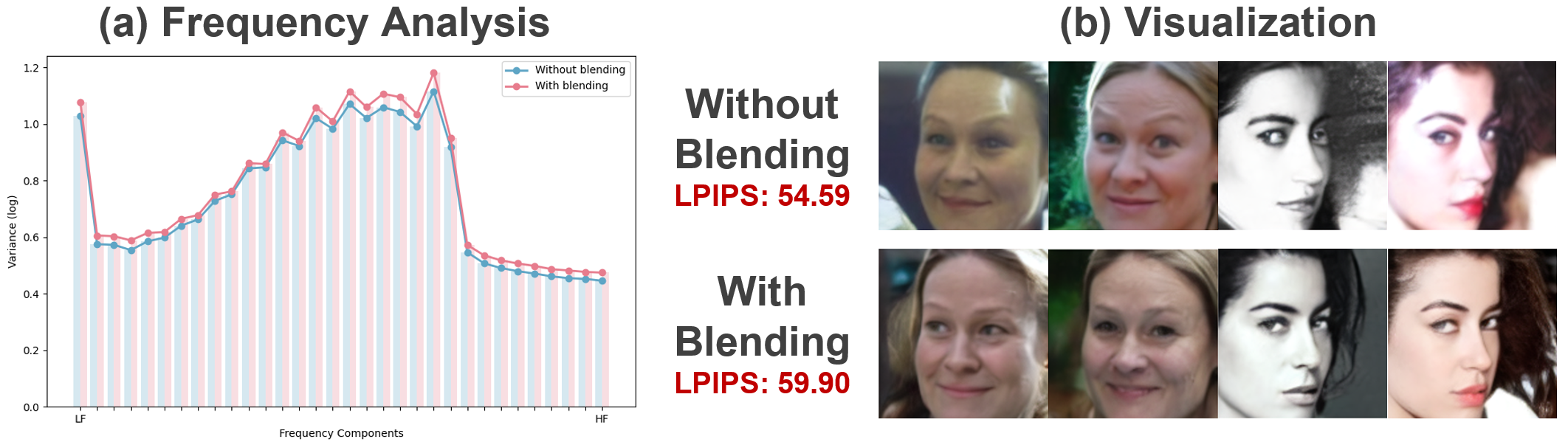}
   \caption{Effects of context blending. It produces images with (a) higher frequency variances and (b) better quality and details.} 
   \label{fig:blending}
   \vspace{-5mm}
\end{figure}

\textit{How does context blending affect performance?} We demonstrate that it mutually benefits intra-class identity consistency and style variation. We compare 4 settings: (1) \textbf{Without blending}, where the generator's denoising is not adjusted with CFG; (2) \textbf{With identity}, where CFG is applied only to the identity context during $[0,t_0]$ timesteps; (3) \textbf{With style}, where CFG only promotes style during $(t_0,T]$; (4) \textbf{With blending}, our advocated setting. 

\noindent \textbf{Quantitative analysis.} Comparisons of eIR, cosine similarity, and FR efficacy are shown in~\cref{tab:exp-sample}(b). We observe: (1) Context blending improves both eIR and cosine similarity, suggesting our approach's effectiveness; (2) Applying CFG to just one context results in either degraded eIR or cosine similarity, and both settings perform slightly worse than without blending, revealing the inherent trade-off between consistency and variation.

\noindent \textbf{Frequency analysis.} We further inspect the frequency components of synthetic images. We convert images into the frequency domain using the fast Fourier transform (FFT) and partition the spectrum into components with different frequencies.  \Cref{fig:blending}(a) shows the dataset-average variances of components. Our proposed setting achieves both higher LF and HF variances compared to without blending, suggesting (though not definitively) more informative styles and identities, respectively.

\noindent \textbf{Visualization.} We compare synthetic images with and without blending in~\cref{fig:blending}(b). We find better diversity (by learned perceptual image patch similarity, LPIPS~\cite{zhang2018unreasonable}) and more facial details (\eg, wrinkles) in with-blending images.
% (1) Using context blending achieves both better eIR and cosine similarity. It promotes average FR performance by 0.21 compared to without blending, suggesting our method's effectiveness; (2) CFGs on one context experiences either degraded eIR or cosine similarity, and both settings performs slightly worse than without blending. This also reveals the inherent trade-off between identity consistency and style variation.

% We further demonstrate that context blending enriches identity and style by frequency analyses. As shown in {\color{blue}Fig. X(a)}, we convert images into frequency spectrum via the fast Fourier transform (FFT) and partition the spectrum into different frequency components. We depict the dataset-wise variances of components in {\color{blue}Fig. X(b)}. We find that our proposed settings achieves higher LF and HF variances than without blending, which reflects (though not decisively) more informative styles and identities, respectively.

% 报告一下LPIPS，注意到有更好的图像质量。也要报告trade-off，把那个高低频分析的图拿出来用一下，画一个示意图解释它是怎么得到的。注意到尽管我们已经很好了，但是真实数据在高频还有丰富得多的细节，作为open problem。也可以比一下身份分布。

\subsection{Privacy Analysis}
\label{subsec:exp-privacy}

\begin{figure}[tbp]
  \centering
   \includegraphics[width=\linewidth]{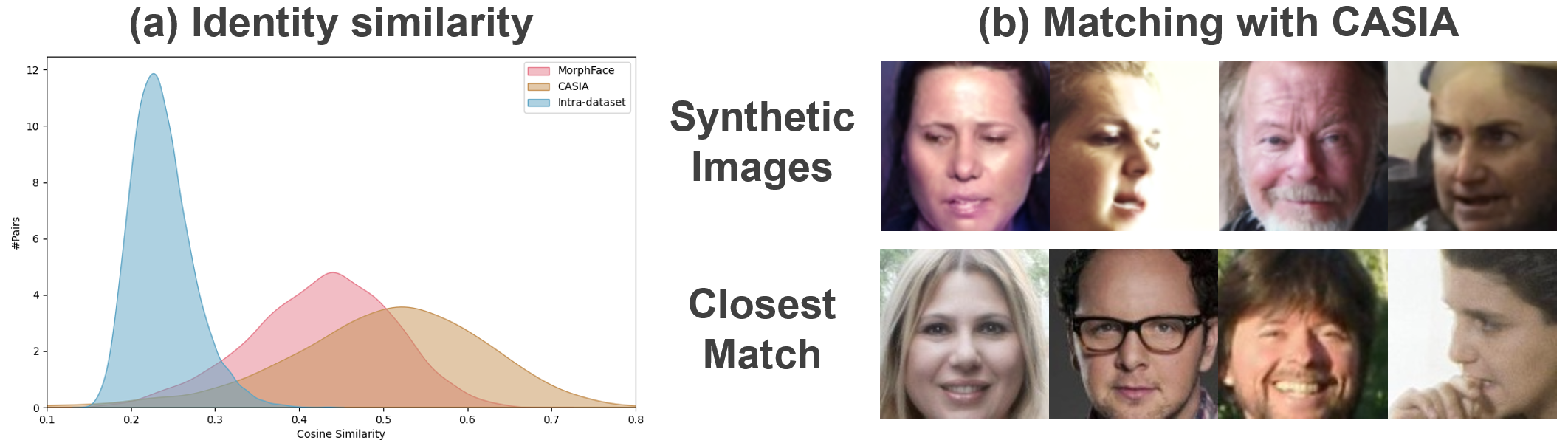}
   \caption{Privacy analyses. (a) Inter-dataset similarity is far lower than the intra-dataset similarities of CASIA and synthetic faces. (b) Synthetic faces are dissimilar to their closest CASIA matches.} 
   \label{fig:privacy}
   \vspace{-3mm}
\end{figure}

The primary purpose of IPFS is to create \textit{unseen} faces that address privacy concerns in real-world datasets. Our generator, $\mathcal{G}$, is trained on CASIA-WebFace~\cite{yi2014learning}, raising the natural question of how similar our synthetic faces are to those in CASIA. High similarity could lead to privacy breaches.

\noindent \textbf{Identity similarity.} CASIA and our synthetic dataset consist of 10.5K and 10K subjects, respectively. We sample one image from each subject and compare the pairwise similarity between subjects from the two datasets. In~\cref{fig:privacy}(a), we compare the inter-dataset similarity with intra-class similarities within each dataset. Both CASIA and our dataset show good intra-class similarities (\ie, 0.52 and 0.45). However, the similarity between them is relatively low (\ie, 0.24). This suggests that our synthetic faces represent virtual subjects, not directly from the training dataset.

\noindent \textbf{Visualization.} In~\cref{fig:privacy}(b), we show sample images from our dataset alongside their closest matches from CASIA. The visual dissimilarity further demonstrates the privacy-preserving nature of our synthetic dataset.

% 缺一个安全分析，拉相似度曲线说明我们的图和CASIA没有身份重合，可以实现安全性；再搞一个top 5查找

\subsection{Ablation Studies}
\label{subsec:exp-abal}

\begin{small}
\begin{table}[tbp]
\centering
\begin{tabular}{lcccc}
\toprule
\textbf{Source}   & \textbf{CFP-FP} & \textbf{AgeDB} & \textbf{CPLFW} & \textbf{CALFW} \\
\midrule
Real-world        & 94.79           & 92.37          & 89.73          & 92.82          \\
Learned     & 91.61           & 91.52          & 85.97          & 92.32          \\
\textbf{Proposed} & 94.11           & 91.80           & 88.73          & 92.73         \\
\bottomrule
\end{tabular}
\caption{Performance on alternative sources of style attributes.}
\label{tab:exp-style-src}
\vspace{-3mm}
\end{table}
\end{small}

\textbf{Alternative sources of style attributes.} We proposed using statistically sampled style attributes. We further compare it with (1) \textbf{Real-world} attributes sampled from CASIA, which represent the theoretical upper-bound performance of our style control; (2) \textbf{Learned} attributes, where we train a VAE on $\mathbf{P}$ to predict $\mathbf{P}'$. From~\cref{tab:exp-style-src}, we infer that: (1) Real-world attributes achieve better performance (partly due to its nature as $\mathcal{G}$'s training data), suggesting potential for future improvements. We note this setting is aligned with~\cite{kim2023dcface} that uses a real style bank; (2) Model-learned attributes perform less well, as they fail to capture the vital statistical details and subject distinctiveness.

% We provide further studies in supplementary material.

% \noindent \textbf{Alternative methods for style conditioning.} We proposed conditioning style from 3DMM renderings using cross-attention. We study 2 alternatives: (1) Directly using style attributes $\mathbf{p}$ as $\mathbf{c}_{sty}$, and (2) Concatenating style feature maps $\mathbf{m}$ to the image channels. From {\color{blue}Fig. X}, we observe that the first approach provides ineffective style control due to a lack of pixel-aligned details. The second approach is incompatible with CFG and may introduce artifacts.
 %, due to space limit.

%% file: sec/5_conclusion.tex
\section{Conclusion}
\label{sec:conclusion}

We have presented MorphFace, a diffusion-based generator that synthesizes faces with both consistent identities and diverse styles. Its advancements are three-fold: (1) Achieving fine-grained, parametric control of facial styles; (2) Creating more realistic style varations that promotes FR efficacy; (3) Enhancing expressiveness of identity and style contexts.

% It learns fine-grained facial styles from the renderings of DECA 3DMM, and learns identity representations from an off-the-shelf recognition model. By conditioning on novel identities of unlabeled synthetic faces and sampling styles from a real-world prior distribution, MorphFace effectively captures both intra-subject variation and subject distinctiveness. Additionally, our proposed context blending optimizes the generator's responsiveness to both identity and style conditions. Extensive experiments have demonstrated the significant performance of MorphFace.

%% file: sec/X_suppl.tex
\clearpage
\setcounter{page}{1}
\setcounter{section}{0}
\renewcommand{\thesection}{\Alph{section}} 
\maketitlesupplementary

\noindent This supplementary material provides more methodological and experimental details that were streamlined in the main text due to space limitations, which we hope is of interest to our readers. It mainly includes:

\begin{enumerate}[label=(\Alph*)]
    \item Detailed experimental setup;
    \item Methodological details; % , including an explanation of the DECA rendering pipeline, a discussion on the multiplicative Gaussian distribution of real-world style attributes, and 
    \item Further explanations to metrics used in the main text; % \ie, cosine similarity, eIR and DECA attributes' variance;
    \item Additional experimental studies;
    \item Visualizations;
    \item Miscellaneous.
\end{enumerate}
% 补充材料：
% 1. 详细的实验设定
% 2. 解释一下实验的细节：DECA的渲染管线，类内是怎样采样的
% 3. 各指标的详细计算方法和对图1/9/10的进一步解释；各指标的细节图；
% 4. 更多消融实验
% 5. 解释乘法高斯分布的原理和实验情况（用智舟之前拉的分布图）
% 6. 关于Idiff（6和7可以开一章Miscellaneous）
% 7. 伦理讨论和代码

\section{Detailed Experimental Setup}
\label{sec:supp-exp-setting}

\subsection{Implementation Details}

We use publicly released FR model $\mathcal{F}$ from ElasticFace~\cite{boutros2022elastic}, unconditional DM $\mathcal{G}_{id}$ from DCFace~\cite{kim2023dcface}, and encoder and decoder $\mathcal{\phi}_e,\mathcal{\phi}_d$ from the VAE of LDM~\cite{rombach2022high}. For the style encoder $\mathcal{E}$, we employ a simple network as depicted in~\cref{fig:supp-pipeline-detail}. We train our generator $\mathcal{G}$ for 250K steps, using an Adam optimizer~\cite{kingma2014adam}, an initial learning rate of 1e-4, and a total batch size of 512.  To incorporate context blending, during training, we replace $\mathbf{c}_{id}$ and $\mathbf{c}_{sty}$ with learnable empty contexts $\mathbf{c}_{id}^{\emptyset}$ and $\mathbf{c}_{sty}^{\emptyset}$ with a probability of 0.1; during inference time, we employ CFG by choosing $t_0$=500 and $w$=0.5. To evaluate our 0.5/1.2M synthetic datasets, we train an IR-50~\cite{duta2021improved} FR model $\mathcal{F}_{syn}$ for 40 epochs using an SGD optimizer~\cite{ruder2016overview}, an ArcFace~\cite{deng2019arcface} loss, a total batch size of 256, and an initial learning rate of 0.1. We employ random horizontal flipping as following the \textit{de facto} standard in FR, and random cropping with a probability of 0.2 as recommended by~\cite{kim2022adaface}. We do not use other forms of data augmentation. We run all experiments on 8 NVIDIA RTX 3090 GPUs and use fixed random seed across all experiments.

\subsection{Datasets}

We train our LDM $\mathcal{G}$ on CASIA-WebFace~\cite{yi2014learning}, a dataset that consists of 490k face images of varied qualities from 10575 identities. We benchmark FR model $\mathcal{F}_{syn}$ trained on our synthetic images on 5 widely used test datasets, LFW~\cite{lfwtechupdate}, CFP-FP~\cite{sengupta2016frontal}, AgeDB~\cite{moschoglou2017agedb}, CPLFW~\cite{zheng2018cross}, and CALFW~\cite{zheng2017cross}. CFP-FP and CPLFW are designed to measure the FR in cross-pose variations, and AgeDB and CALFW are for cross-age variations.

\subsection{Critical Feature Shapes}

$\mathcal{G},\mathcal{G}_{id}$ produces 3×128×128 images. Each of the 3DMM feature maps (surface normals, albedo, Lambertian rendering) is 3×128×128, and their concatenation by channel is 9×128×128. The latent representation of $\mathcal{G}$ is 3×32×32. For the training of FR model $\mathcal{F}_{syn}$, we resize the synthetic images into 3×112×112 to match $\mathcal{F}_{syn}$'s input shape. The lengths of $\mathbf{c}_{id}$ and $\mathbf{c}_{sty}$ are 512.

\section{Methodological Details}
\label{sec:supp-method}

\begin{figure*}[tbp]
  \centering
   \includegraphics[width=\linewidth]{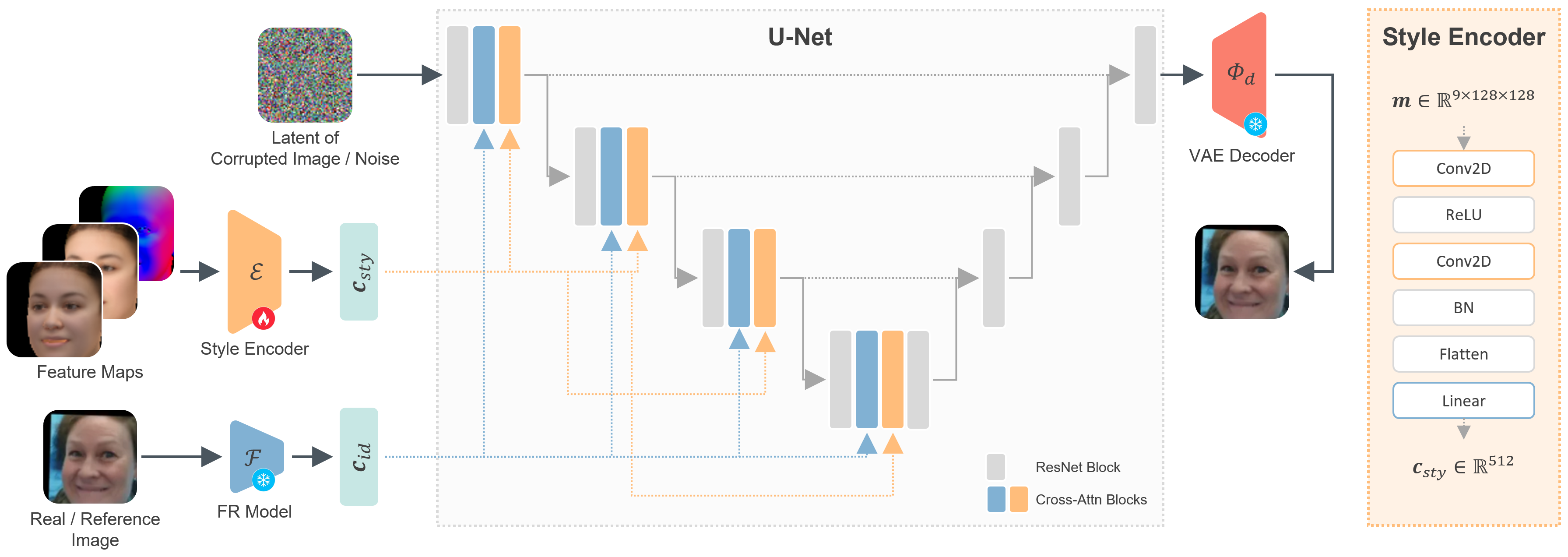}
   \caption{A detailed look at MorphFace generator $\mathcal{G}$ and style encoder $\mathcal{E}$. The main body of the generator is a U-Net~\cite{ronneberger2015u} noise estimator. The identity and style contexts $\mathbf{c}_{id},\mathbf{c}_{sty}$ are incorporated into the model via cross-attention layers after the U-Net's ResNet blocks. By cross-attention, we follow the same practice described in Sec. 3.3 of the LDM fundamental paper~\cite{rombach2022high}. The style encoder $\mathcal{E}$ is a rather simple module consisting of 2 convolution layers plus a linear layer.} 
   \label{fig:supp-pipeline-detail}
\end{figure*}

\subsection{Style Extraction}

Our proposed approach uses an off-the-shelf DECA 3DMM $\mathcal{M}$ to extract style attributes $\mathbf{p}$ and render them into feature maps $\mathbf{m}$. We briefly digest these attributes and feature maps from the DECA paper~\cite{feng2021learning} to help explain their details.

\noindent \textbf{Style attribute extraction.} Given input face image $\mathbf{x}\in\mathbb{R}^{3\times128\times128}$, DECA uses a trained encoder to infer 6 attribute groups that entirely describe the face's style: (1) \textbf{Shape} $p_s\in \mathbb{R}^{100}$, representing facial geometry features decomposed via principal component analysis (PCA). Each dimension controls a specific geometric aspect, \eg, the width of facial contours; (2) \textbf{Expression} $p_e\in \mathbb{R}^{50}$, describing facial expression features extracted through PCA; (3\&4) \textbf{Pose and Camera} $p_p\in\mathbb{R}^{9}$. Pose is represented in 3D coordinates, while the camera models the projection from the 3D facial mesh to 2D space. Since the image's pose is jointly determined by both 3D pose and camera information, we collectively refer to them as ``pose'' for simplicity; (5) \textbf{Texture} $p_t\in \mathbb{R}^{50}$, modeling facial textures such as wrinkles, derived via PCA; (6) \textbf{Illumination} $p_i\in \mathbb{R}^{27}$, describing lighting conditions on the facial 3D mesh using spherical harmonics. For simplicity, we represent these attributes together as a unified style vector $\mathbf{p}\in \mathbb{R}^{236}$ in our main text.

\noindent \textbf{Feature map rending.} DECA renders 3 feature maps based on the extracted style attributes. First, it generates a 3D facial mesh using FLAME~\cite{li2017learning}, combining shape, expression, and pose attributes. The 3D mesh contains 5023 vertices. It then renders the mesh into the following feature maps: (1) \textbf{Surface Normals} $m_s\in\mathbb{R}^{3\times128\times128}$, representing facial geometry as the normal vectors of each vertex in the mesh; (2) \textbf{Albedo} $m_a\in\mathbb{R}^{3\times128\times128}$, capturing facial texture without lighting effects, derived by combining the mesh with texture attributes;
(3) \textbf{Lambertian Rendering} $m_l\in\mathbb{R}^{3\times128\times128}$, a coarse rendering that incorporates both texture and illumination attributes. These three feature maps provide a detailed description of facial styles and are concatenated along the channel dimension into $\mathbf{m}\in \mathbb{R}^{9\times128\times128}$. This consolidated representation effectively supports style control.

% \noindent \textbf{Correction.} We correct a minor mistake in our pipeline figure (\cref{fig:pipeline}): The ``3D mesh'' should appear after ``style attributes'', as part of DECA's rendering process.

\subsection{Architecture of LDM Generator}

\Cref{fig:supp-pipeline-detail} provides a detailed look at MorphFace generator $\mathcal{G}$ and style encoder $\mathcal{E}$. The generator's main body is a U-Net~\cite{ronneberger2015u} noise estimator. The identity and style contexts $\mathbf{c}_{id},\mathbf{c}_{sty}$ are incorporated into the model via cross-attention layers after the U-Net's ResNet blocks. By cross-attention, we follow the same practice described in Sec. 3.3 of the LDM paper~\cite{rombach2022high}. The style encoder $\mathcal{E}$ is a simple module including 2 convolution layers plus a linear layer.

\subsection{Distribution of Style Attributes}

\begin{figure}[tbp]
  \centering
   \includegraphics[width=\linewidth]{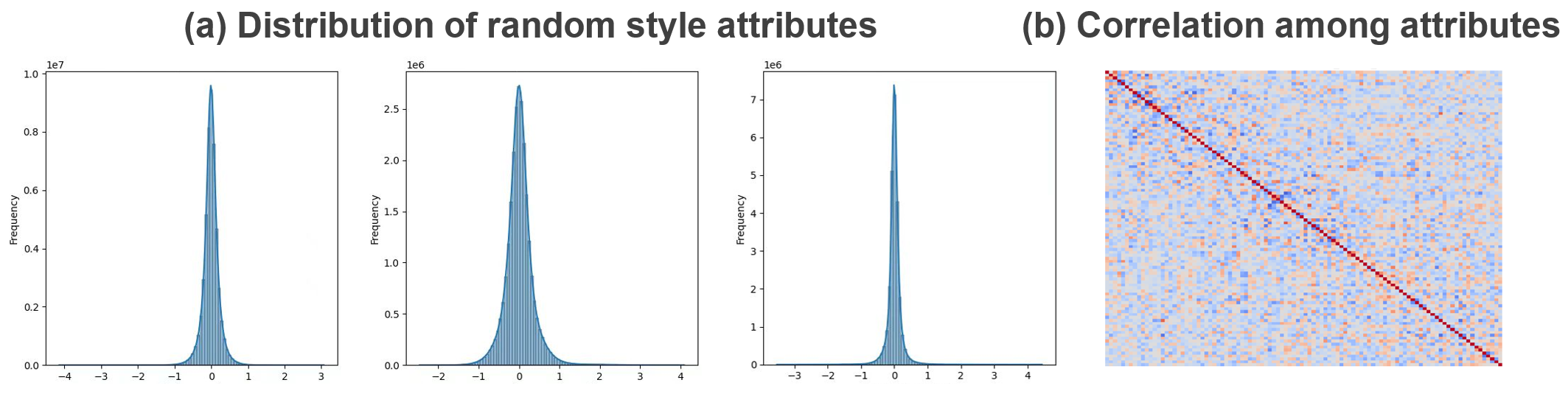}
   \caption{Distribution of style attributes. (a) We find Gaussian distributions in randomly chosen attribute dimensions from $\mathbf{P}$. (b) Sample correlation matrix of $\mathbf{P}$'s shape dimensions.} 
   \label{fig:supp-attribute-dist}
\end{figure}

In~\cref{subsec:method-sample}, we approximate the distribution of real-world style attributes by a multiplicative Gaussian distribution, \ie, $\mathbb{D}(\mathbf{P})$$\sim$$\mathcal{N}(\mathbf{\mu},\mathbf{\Sigma})$. We here explain its rationale: (1) Each attribute dimension of shape, expression and texture follows a Gaussian distribution as a natural outcome of DECA~\cite{feng2021learning}. In DECA, these attributes are derived from PCA, and Gaussian distribution is part of PCA's assumption. (2) Previous findings~\cite{paysan20093d,booth2018large} suggest that facial attributes including pose and illumination can be modeled via Gaussian distributions. As each attribute dimension can be considered as an approximation of Gaussian distribution, their multiplication holds $\mathbb{D}(\mathbf{P})$$\sim$$\mathcal{N}(\mathbf{\mu},\mathbf{\Sigma})$.

We also empirically validate the assumption. We find Gaussian distributions in randomly chosen attribute dimensions from $\mathbf{P}$, as shown in~\cref{fig:supp-attribute-dist}(a). Here, we can also infer each dimension's mean $\mu_i$ and variance $\epsilon_i$. In~\cref{fig:supp-attribute-dist}(b), we visualize the correlation matrix of $\mathbf{P}$'s shape dimensions. Knowing each dimension's mean and variance, and the correlation matrix allows concretizing $\mathcal{N}(\mathbf{\mu},\mathbf{\Sigma})$.

\subsection{Classifier-Free Guidance}

In~\cref{subsec:method-blend}, we employ CFG~\cite{ho2022classifier} for context blending. CFG is a common technique in generative models, particularly DMs, to strengthen the generated samples' adherence to conditioning contexts without an explicit classifier.

In the training phase, CFG requires the model to be trained to predict the noise added to data for two scenarios, (1) conditional, when conditioning context (\ie, $\mathbf{c}_{id}$ and $\mathbf{c}_{sty}$ in our case) is provided, and (2) unconditional, when the context is null or a placeholder. We achieve unconditional training by probabilistically replacing $\mathbf{c}_{id}$ and $\mathbf{c}_{sty}$ with learnable empty contexts $\mathbf{c}_{id}^{\emptyset}$ and $\mathbf{c}_{sty}^{\emptyset}$, \ie, the placeholders. In the inference phase, the predicted noise $\mathbf{\epsilon}_{cfg}$ is computed as a weighted combination of conditional and unconditional predictions as

\begin{equation}
    \mathbf{\epsilon}_{cfg}=(1+w)\mathbf{\epsilon}_{\theta}(\mathbf{z}_t, t, \mathbf{c}) - w\mathbf{\epsilon}_{\theta}(\mathbf{z}_t, t, \mathbf{c}^{\emptyset}),
\end{equation}

\noindent where $w$$>$0 strengthens the condition. We concretize $\mathbf{\epsilon}_{\theta}(\mathbf{z}_t, t, \mathbf{c}^{\emptyset})$ as $\mathbf{\epsilon}_{\theta}(\mathbf{z}_t, t, \mathbf{c}_{id},\mathbf{c}_{sty}^{\emptyset})$ and $\mathbf{\epsilon}_{\theta}(\mathbf{z}_t, t, \mathbf{c}_{id}^{\emptyset},\mathbf{c}_{sty})$ to incorporate dual conditions, to augment style and identity, respectively.

\section{Metrics Explained}
\label{sec:supp-metric}

\begin{figure}[tbp]
  \centering
   \includegraphics[width=\linewidth]{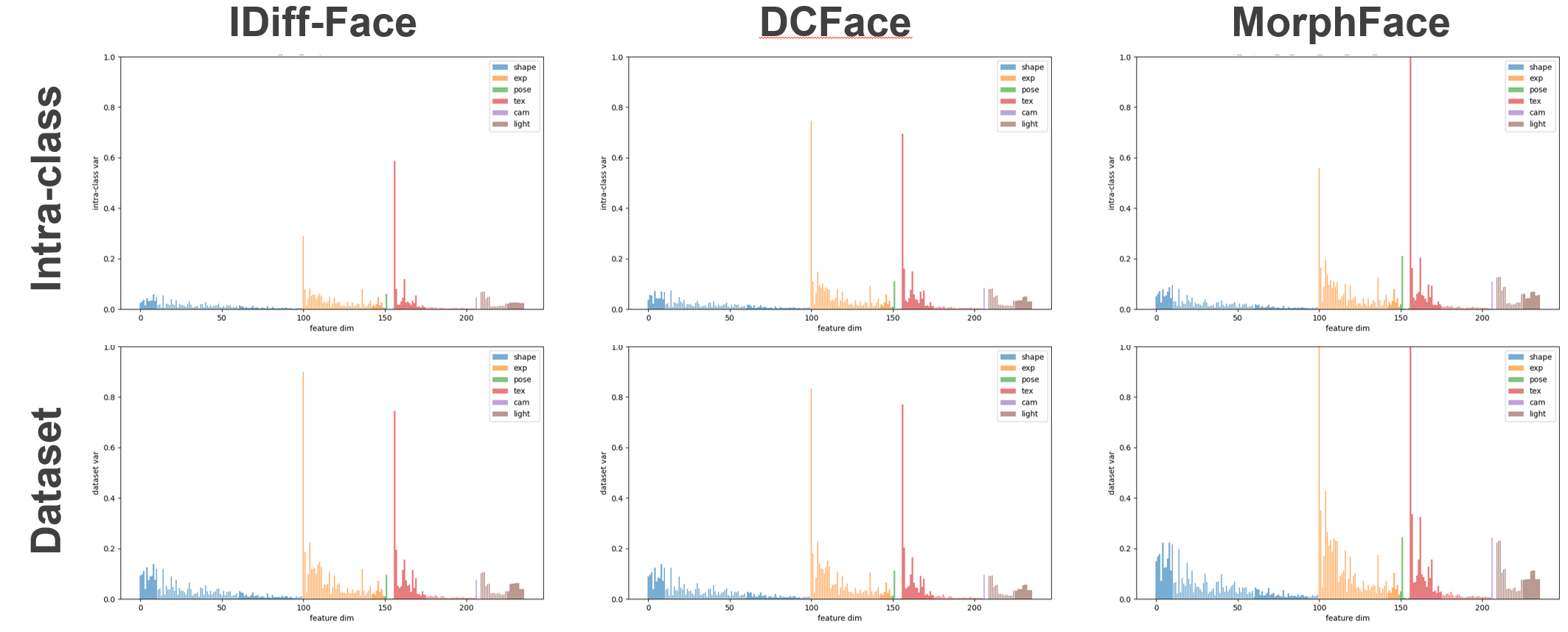}
   \caption{Variances of DECA-extracted style attributes. The same result is streamlined in~\cref{fig:paradigm}. Larger intra-class and dataset-wise variance represent better intra-class style variation and dataset variability. It can be inferred that IDiff-Face is inadequate in style variation and MorphFace has diverse varied styles.} 
   \label{fig:supp-style-var}
\end{figure}

We explain the details of 4 metrics we used in the main text.

\subsection{Cosine Similarity}

It is the fundamental metric in SOTA FR systems to measure the similarity between two identity embeddings representing face images. Formally, let $\textbf{x}_1,\textbf{x}_2$ denote two face images, $\mathcal{F}$ denote a pre-trained FR model, and $\textbf{d}_1,\textbf{d}_2$ denote the identity embeddings extracted as $\textbf{d}=\mathcal{F}(\mathbf{x})$. $\textbf{d}_1,\textbf{d}_2$ are 512-dim feature vectors in our case. The cosine similarity between $\textbf{d}_1,\textbf{d}_2$ is

\begin{equation}
    \operatorname{cossim}(\textbf{d}_1,\textbf{d}_2) = \frac{\textbf{d}_1\cdot\textbf{d}_2}{|\textbf{d}_1||\textbf{d}_2|}.
\end{equation}

\noindent A larger $\operatorname{cossim}(\textbf{d}_1,\textbf{d}_2)$ indicates that $\textbf{x}_1,\textbf{x}_2$ are more likely the same person. To train an effective FR model, we expect the training face images to have high intra-class cosine similarity (\ie, identity consistency within each subject) and low inter-class cosine similarity (\ie, unique subjects). \textit{In our main text:} (1) In~\cref{fig:paradigm}, we depict curves of intra-class and inter-class cosine similarities, hence more separated curves indicate better FR efficacy. (2) In~\cref{fig:trade-off} and~\cref{tab:exp-sample}, we report the average intra-class cosine similarity. (3) In~\cref{subsec:method-sample}, we only enroll reference images with cosine similarity below 0.3 to filter those less distinct subjects.

\subsection{DECA Attribute Variance}

In~\cref{subsec:method-train}, we use a pre-trained DECA 3DMM to infer the style attributes from an input image, $\mathbf{p}$$=$$\mathcal{M}(\mathbf{x})$. The style attributes can be considered a 236-dim vector, where its 100, 50, 9, 50, and 27 dimensions represent the image $\mathbf{x}$'s facial shape, expression, pose, texture, and illumination, respectively. As the image's style is solely parameterized by style attributes, the intra-class and dataset-wise variances of these attributes demonstrate the dataset's intra-class style variation and dataset variability. \textit{In our main text:} In~\cref{fig:paradigm}, we depict the average style variance of facial shape, expression, pose, texture, and illumination, hence larger shaded areas represent better intra-class style variation and dataset variability. We supplement the detailed attribute-wise variance in~\cref{fig:supp-style-var}. It can be inferred that IDiff-Face is inadequate in style variation and MorphFace has diverse styles.

\subsection{Extended Improved Recall (eIR)}

It is proposed by DCFace~\cite{kim2023dcface} as an extension of Improved Recall~\cite{kynkaanniemi2019improved} to measure the style diversity of synthetic images. The images $\mathbf{x}$ are first mapped into a style latent space via an Inception Network~\cite{salimans2016improved} trained on ImageNet~\cite{deng2009imagenet} to obtain inception vectors $\mathbf{v}$. To calculate eIR, for a set of real (\ie, CASIA) and synthetic inception vectors $\{\mathbf{v}_i^c\},\{\mathbf{\hat{v}}_{j}^{c}\}$ under the same label condition $c$, define the $k$-nearest feature distance $r_k$ as $r_k$$=$$d(\mathbf{\hat{v}}_{j}^{c} - \operatorname{NN}_k(\mathbf{\hat{v}}_{j}^{c},\{\mathbf{\hat{v}}_{j}^{c}\})$ where $\operatorname{NN}_k$ returns the $k$-nearest vectors in $\{\mathbf{\hat{v}}_{j}^{c}\}$ and 

\begin{equation}
    \mathbf{I}(\mathbf{v}_i^c,\{\mathbf{\hat{v}}_{j}^{c}\})=
    \begin{cases}
        1,\quad \exists \mathbf{\hat{v}}_{j}^{c}\in\{\mathbf{\hat{v}}_{j}^{c}\}~\text{s.t.}~d(\mathbf{v}_i^c, \mathbf{\hat{v}}_{j}^{c}) < r_k, \\
        0,\quad \text{otherwise},
    \end{cases}
\end{equation}

\noindent $d(\cdot)$ is $l_2$ distance. The eIR is defined as 

\begin{equation}
    \operatorname{eIR} = \frac{1}{C}\frac{1}{\sum_c N_c} \sum_{c=1}^{C} \sum_{i=1}^{N_c} \left(\mathbf{I}(\mathbf{v}_i^c,\{\mathbf{\hat{v}}_{j}^{c}\})\right),
\end{equation}

\noindent which is the fraction of real image styles manifold covered by the synthetic image style manifold as defined by $k$-nearest neighbor ball. If the style variation is small, then $r_k$ becomes small, reducing the chance of $d(\mathbf{v}_i^c, \mathbf{\hat{v}}_{j}^{c}) < r_k$. \textit{In our main text:} (1) In~\cref{fig:trade-off}, we measure intra-class style variation by intra-class eIR and dataset variability by dataset eIR. (2) In~\cref{tab:exp-style-src}, we report intra-class eIR for each setting.

\begin{figure}[tbp]
  \centering
   \includegraphics[width=\linewidth]{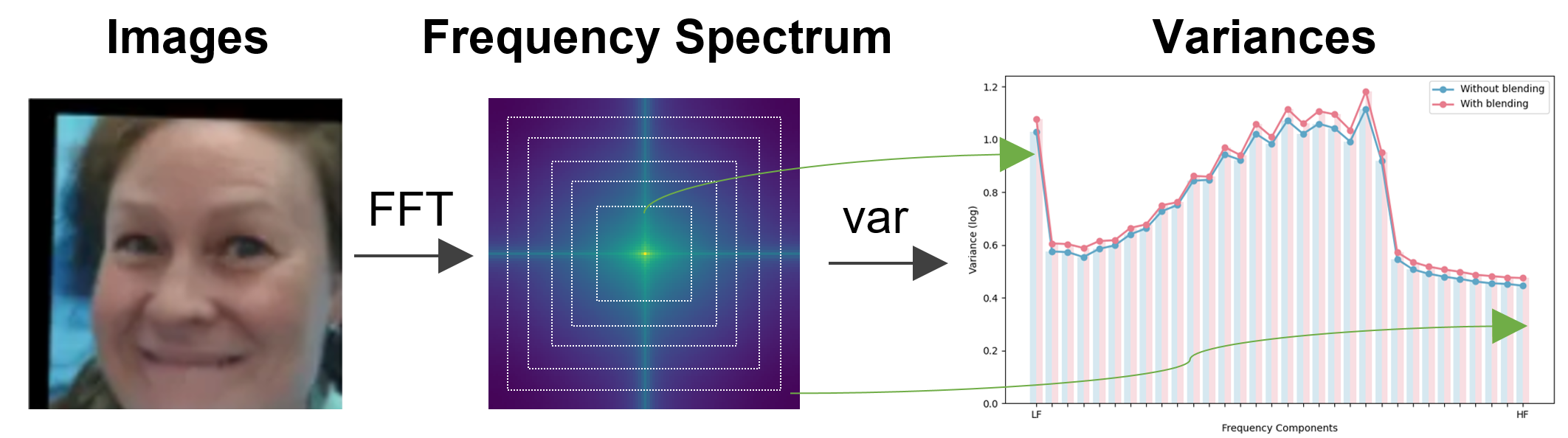}
   \caption{The calculation of frequency variances. Images are converted into frequency spectrum via FFT and partitioned into different frequency components, where their variances are measured. Higher variances reflect more informative frequency components.} 
   \label{fig:supp-freq}
\end{figure}

\subsection{Frequency Variances}

It measures the diversity across different frequency components. \Cref{fig:supp-freq} explains its calculation, where images are converted into frequency spectrum via FFT and partitioned into different frequency components, and the variance of each component is measured. Higher variances reflect (though not in a decisive manner) more informative frequency components, hence better identity consistency and style variation. \textit{In our main text}, it is exhibited in~\cref{fig:blending}(a).

\section{Additional Experimental Studies}
\label{sec:supp-abal}

\subsection{Shape Attribute Replacement}

\begin{figure}[tbp]
  \centering
   \includegraphics[width=\linewidth]{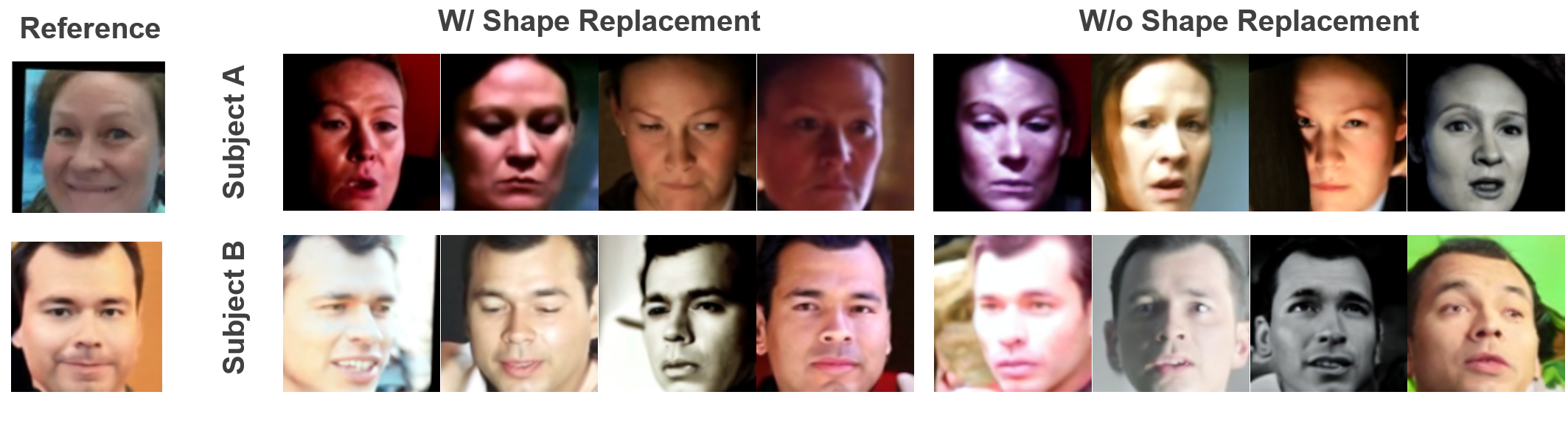}
   \caption{Comparison of synthetic images with and without replacing their shape attributes with ground-truth shape during style sampling. Shape replacement offers synthetic images with improved visual similarity to the reference image.} 
   \label{fig:supp-shape-replace}
\end{figure}

As discussed in~\cref{subsec:method-sample}, we sample style attributes (\ie, facial shape, expression, pose, texture and illumination) from a real-world prior distribution. Then, we replace the intra-class mean of facial shape attributes with the reference image's ground-truth shape. In~\cref{fig:supp-shape-replace}, we compare synthetic images with and without replacing their shape attributes during style sampling. Shape replacement offers synthetic images with better visual similarity to the reference image. Experimentally, this improves average FR accuracy by 0.22.

\begin{figure}[tbp]
  \centering
   \includegraphics[width=\linewidth]{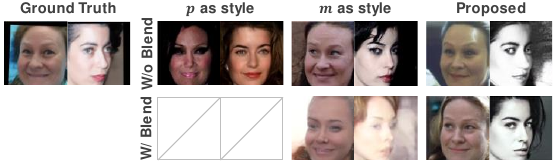}
   \caption{Alternative methods for style conditioning. Directly using style attributes as context fails to control style due to lacking pixel-aligned details. Concatenating style feature maps to image channels produces artifacts when incorporating blending.} 
   \label{fig:supp-alter-style}
\end{figure}

\subsection{Alternation for Style Conditioning}

 We proposed to condition style from 3DMM renderings using cross-attention. We study 2 alternatives: (1) Directly using style attributes $\mathbf{p}$ as $\mathbf{c}_{sty}$, and (2) Concatenating style feature maps $\mathbf{m}$ to the image channels. From~\cref{fig:supp-alter-style}, we observe that the first approach provides ineffective style control due to a lack of pixel-aligned details. Though the second approach controls style, we find it incompatible with context blending (as CFG ineffectively learns empty feature maps) and may introduce artifacts.

\begin{figure}[tbp]
  \centering
   \includegraphics[width=\linewidth]{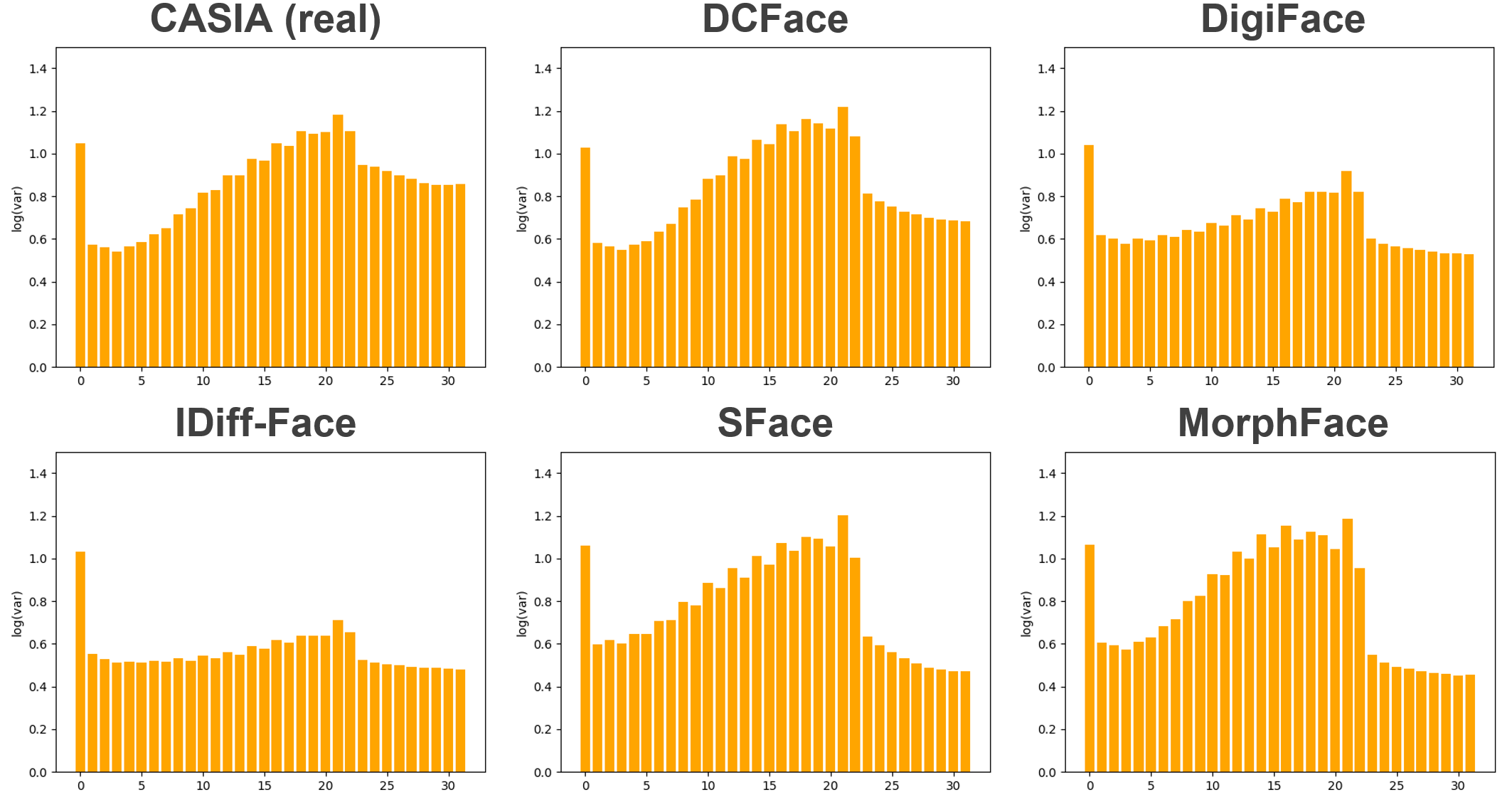}
   \caption{Comparison with SOTAs on frequency variances. Our proposed method, DCFace and SFace exhibit informative frequency components similar to CASIA. The analyses match the quantitative eIR results in~\cref{fig:trade-off}(a).} 
   \label{fig:supp-freq-sota}
\end{figure}

\subsection{Comparison on Frequency Variance}

In~\cref{fig:supp-freq-sota}, we measure the intra-class variances of frequency components for the real-world CASIA dataset, several SOTAs, and our proposed MorphFace. This is an extension of~\cref{fig:blending}(a) of our main text. We highlight: (1) MorphFace, DCFace and SFace exhibit informative frequency components similar to CASIA, while DigiFace and SFace exhibit less informative components. (2) The frequency analyses match the quantitative eIR results in~\cref{fig:trade-off}(a), where MorphFace, DCFace and SFace have higher intra-class eIR. This also demonstrates the reasonableness of frequency analyses.

\begin{small}
    \begin{table}[tbp]
    \centering
\begin{tabular}{llccc}
\toprule
& \textbf{Strategy}      & \textbf{eIR} & \textbf{cos-sim} & \textbf{FR Avg.} \\

\midrule
& W/o blending           & 0.608        & 0.37             & 93.11            \\
\midrule
\multirow{4}{*}
{\begin{tabular}[c]{@{}l@{}}\textbf{$t_0$} \end{tabular}} 
& 750           & 0.617        & \textbf{0.48}             & 93.23            \\
& 250                & \textbf{0.675}        & 0.38             & 93.18            \\
& \textbf{500} (Proposed) & 0.642        & 0.45             & \textbf{93.32}            \\
\midrule
\multirow{4}{*}{\begin{tabular}[c]{@{}l@{}}\textbf{$w$}\end{tabular}}
& 1           & \textbf{0.684}        & \textbf{0.51}             & 93.05            \\
& 0.25           & 0.613        & 0.37             & 93.14            \\
& \textbf{0.5} (Proposed)   & 0.642        & 0.45             & \textbf{93.32}            \\
\bottomrule
\end{tabular}
    \caption{Choices of shifting timesteps $t_0$ and CFG weight $w$.}
    \label{tab:supp-blend}
    \end{table}
\end{small}

\begin{figure}[tbp]
  \centering
   \includegraphics[width=\linewidth]{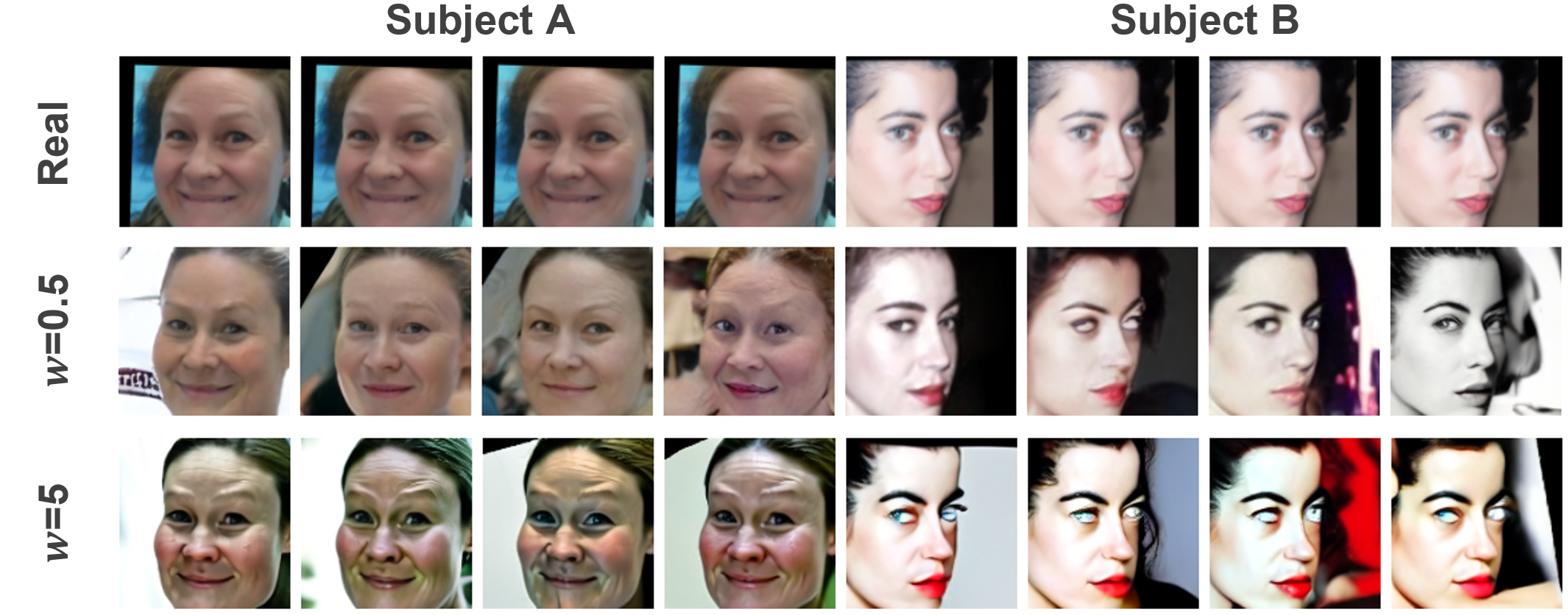}
   \caption{Sample images from different CFG weight $w$. A too-intensive weight (\eg, 5) could produce less realistic images that downgrade FR efficacy.} 
   \label{fig:supp-abal-blend}
\end{figure}

\subsection{Choice of Blending Parameters}

We study the impact of choosing different shifting timesteps $t_0$ and CFG weight $w$ during context blending on synthesizing quality and FR efficacy. Results are summarized by eIR, cosine similarity and average FR accuracy in~\cref{tab:supp-blend}. We highlight: (1) Choosing larger/smaller $t_0$ strengthens the impact of identity/style contexts, leading to increased cosine similarity/eIR, respectively. They both suffer a slight accuracy drop, suggesting the importance of balancing between contexts. The drop however is slight and both settings outperform the non-blending baseline, demonstrating the effectiveness of our proposed technique. (2) By choosing a smaller $w$=0.25, context blending is too inconspicuous to affect performance. (3) Choosing a larger $w$=1 increases both eIR and cosine similarity. Interestingly, this negatively impacts FR efficacy. In~\cref{fig:supp-abal-blend}, we find an intensive $w$ (\eg, a very large 5) could generate less realistic images, which explains the accuracy downgrade. This suggests that a moderate $w$ should be chosen for context blending. We leave its improvement in future studies.

\section{Visualizations}
\label{sec:supp-vis}

\begin{figure}[tbp]
  \centering
   \includegraphics[width=\linewidth]{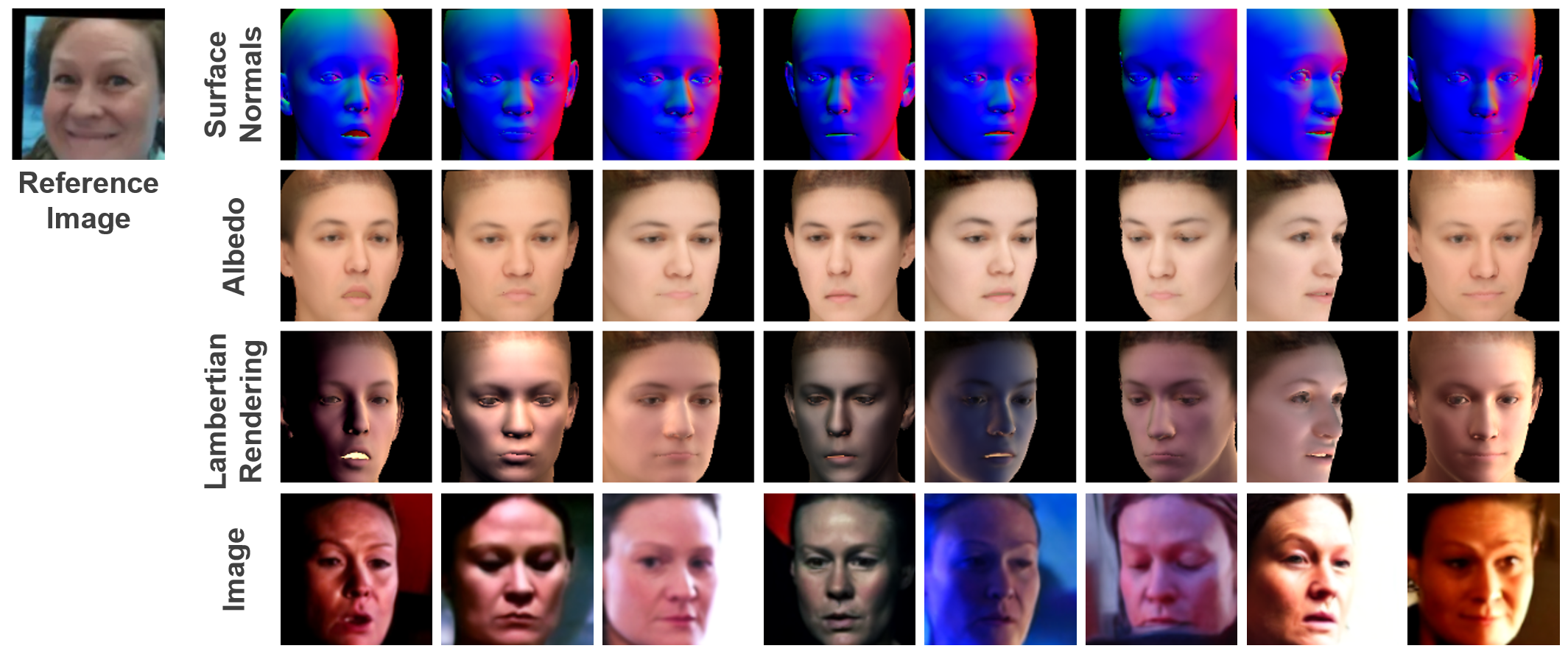}
   \caption{Sample DECA feature maps and their synthetic images.} 
   \label{fig:supp-add-feat}
   \vspace{-2mm}
\end{figure}

\begin{figure}[tbp]
  \centering
   \includegraphics[width=\linewidth]{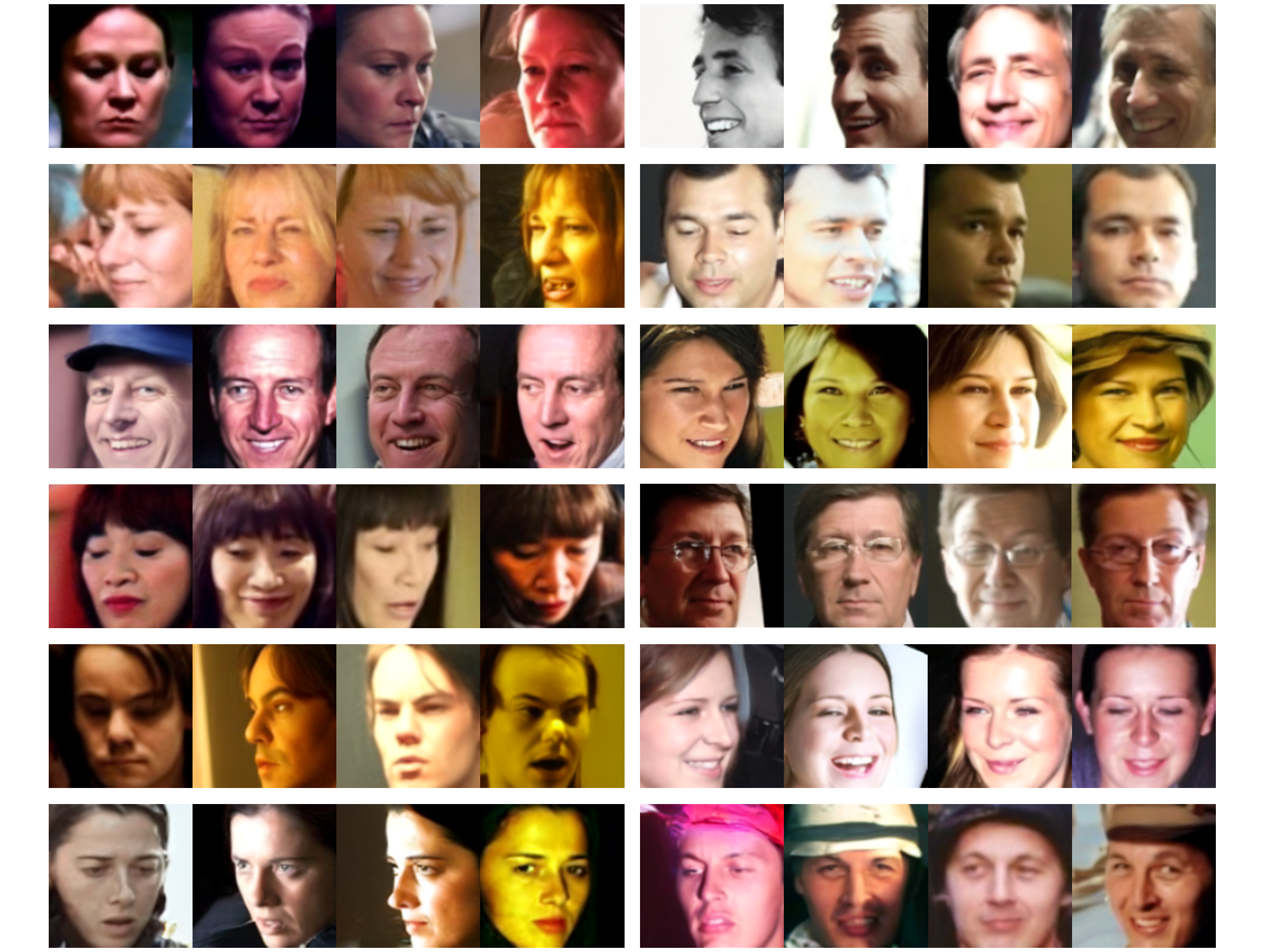}
   \caption{Additional sample images from MorphFace.} 
   \label{fig:supp-add-syn}
\end{figure}

In~\cref{fig:supp-add-feat}, we provide sample DECA feature maps $\mathbf{m}$ and their synthetic images. As discussed in~\cref{subsec:method-train,subsec:method-sample}, the feature maps are rendered from style attributes $\mathbf{p}'$, whose expression, pose, texture and illumination are randomly sampled from a real-world prior distribution and shape comes from the reference image. \Cref{fig:supp-add-feat} is an extension of~\cref{fig:vis-3dmm} in our main text.

In~\cref{fig:supp-add-syn}, we provide additional sample images from MorphFace, where intra-class style variation and subject distinctiveness can both be observed. The 0.5/1.2M synthetic datasets will be later released for public access.

\section{Miscellaneous}
\label{sec:supp-misc}

\noindent \textbf{Code and dataset.} The code and synthetic datasets will be available at \url{https://github.com/Tencent/TFace/}.

% {\color{blue}\noindent \textbf{Societal impacts.}}

% \noindent \textbf{Code and dataset.} We promise to release our code and 0.5/1.2M synthetic datasets promptly after the manuscript's publication. We kindly appreciate the reviewers' patience.

%% file: main.bbl
\begin{thebibliography}{97}
\providecommand{\natexlab}[1]{#1}
\providecommand{\url}[1]{\texttt{#1}}
\expandafter\ifx\csname urlstyle\endcsname\relax
  \providecommand{\doi}[1]{doi: #1}\else
  \providecommand{\doi}{doi: \begingroup \urlstyle{rm}\Url}\fi

\bibitem[Bae et~al.(2023)Bae, de~La~Gorce, Baltru{\v{s}}aitis, Hewitt, Chen, Valentin, Cipolla, and Shen]{bae2023digiface}
Gwangbin Bae, Martin de La~Gorce, Tadas Baltru{\v{s}}aitis, Charlie Hewitt, Dong Chen, Julien Valentin, Roberto Cipolla, and Jingjing Shen.
\newblock Digiface-1m: 1 million digital face images for face recognition.
\newblock In \emph{Proceedings of the IEEE/CVF Winter Conference on Applications of Computer Vision}, pages 3526--3535, 2023.

\bibitem[Blanz and Vetter(2023)]{blanz2023morphable}
Volker Blanz and Thomas Vetter.
\newblock A morphable model for the synthesis of 3d faces.
\newblock \emph{Seminal Graphics Papers: Pushing the Boundaries, Volume 2}, pages 157--164, 2023.

\bibitem[Booth et~al.(2018)Booth, Roussos, Ponniah, Dunaway, and Zafeiriou]{booth2018large}
James Booth, Anastasios Roussos, Allan Ponniah, David Dunaway, and Stefanos Zafeiriou.
\newblock Large scale 3d morphable models.
\newblock \emph{International Journal of Computer Vision}, 126\penalty0 (2):\penalty0 233--254, 2018.

\bibitem[Boutros et~al.(2022{\natexlab{a}})Boutros, Damer, Kirchbuchner, and Kuijper]{boutros2022elastic}
Fadi Boutros, Naser Damer, Florian Kirchbuchner, and Arjan Kuijper.
\newblock Elasticface: Elastic margin loss for deep face recognition.
\newblock In \emph{Proceedings of the IEEE/CVF Conference on Computer Vision and Pattern Recognition (CVPR) Workshops}, pages 1578--1587, 2022{\natexlab{a}}.

\bibitem[Boutros et~al.(2022{\natexlab{b}})Boutros, Huber, Siebke, Rieber, and Damer]{boutros2022sface}
Fadi Boutros, Marco Huber, Patrick Siebke, Tim Rieber, and Naser Damer.
\newblock Sface: Privacy-friendly and accurate face recognition using synthetic data.
\newblock In \emph{2022 IEEE International Joint Conference on Biometrics (IJCB)}, pages 1--11. IEEE, 2022{\natexlab{b}}.

\bibitem[Boutros et~al.(2022{\natexlab{c}})Boutros, Siebke, Klemt, Damer, Kirchbuchner, and Kuijper]{boutros2022pocketnet}
Fadi Boutros, Patrick Siebke, Marcel Klemt, Naser Damer, Florian Kirchbuchner, and Arjan Kuijper.
\newblock Pocketnet: Extreme lightweight face recognition network using neural architecture search and multistep knowledge distillation.
\newblock \emph{IEEE Access}, 10:\penalty0 46823--46833, 2022{\natexlab{c}}.

\bibitem[Boutros et~al.(2023{\natexlab{a}})Boutros, Grebe, Kuijper, and Damer]{boutros2023idiff}
Fadi Boutros, Jonas~Henry Grebe, Arjan Kuijper, and Naser Damer.
\newblock Idiff-face: Synthetic-based face recognition through fizzy identity-conditioned diffusion model.
\newblock In \emph{Proceedings of the IEEE/CVF International Conference on Computer Vision}, pages 19650--19661, 2023{\natexlab{a}}.

\bibitem[Boutros et~al.(2023{\natexlab{b}})Boutros, Klemt, Fang, Kuijper, and Damer]{boutros2023exfacegan}
Fadi Boutros, Marcel Klemt, Meiling Fang, Arjan Kuijper, and Naser Damer.
\newblock Exfacegan: Exploring identity directions in gan’s learned latent space for synthetic identity generation.
\newblock In \emph{2023 IEEE International Joint Conference on Biometrics (IJCB)}, pages 1--10. IEEE, 2023{\natexlab{b}}.

\bibitem[Boutros et~al.(2023{\natexlab{c}})Boutros, Klemt, Fang, Kuijper, and Damer]{boutros2023unsupervised}
Fadi Boutros, Marcel Klemt, Meiling Fang, Arjan Kuijper, and Naser Damer.
\newblock Unsupervised face recognition using unlabeled synthetic data.
\newblock In \emph{2023 IEEE 17th International Conference on Automatic Face and Gesture Recognition (FG)}, pages 1--8. IEEE, 2023{\natexlab{c}}.

\bibitem[Boutros et~al.(2024)Boutros, Huber, Luu, Siebke, and Damer]{boutros2024sface2}
Fadi Boutros, Marco Huber, Anh~Thi Luu, Patrick Siebke, and Naser Damer.
\newblock Sface2: Synthetic-based face recognition with w-space identity-driven sampling.
\newblock \emph{IEEE Transactions on Biometrics, Behavior, and Identity Science}, 2024.

\bibitem[Cao et~al.(2018)Cao, Shen, Xie, Parkhi, and Zisserman]{cao2018vggface2}
Qiong Cao, Li Shen, Weidi Xie, Omkar~M Parkhi, and Andrew Zisserman.
\newblock Vggface2: A dataset for recognising faces across pose and age.
\newblock In \emph{2018 13th IEEE international conference on automatic face \& gesture recognition (FG 2018)}, pages 67--74. IEEE, 2018.

\bibitem[Chen et~al.(2023{\natexlab{a}})Chen, Zhao, Liu, Ding, Song, Wang, Wang, Yang, Liu, Du, et~al.]{chen2023photoverse}
Li Chen, Mengyi Zhao, Yiheng Liu, Mingxu Ding, Yangyang Song, Shizun Wang, Xu Wang, Hao Yang, Jing Liu, Kang Du, et~al.
\newblock Photoverse: Tuning-free image customization with text-to-image diffusion models.
\newblock \emph{arXiv preprint arXiv:2309.05793}, 2023{\natexlab{a}}.

\bibitem[Chen et~al.(2023{\natexlab{b}})Chen, Fang, Liu, He, Huang, Zhang, and Mao]{chen2023dreamidentity}
Zhuowei Chen, Shancheng Fang, Wei Liu, Qian He, Mengqi Huang, Yongdong Zhang, and Zhendong Mao.
\newblock Dreamidentity: Improved editability for efficient face-identity preserved image generation.
\newblock \emph{arXiv preprint arXiv:2307.00300}, 2023{\natexlab{b}}.

\bibitem[Chen et~al.(2024)Chen, Lu, Chen, Song, and Li]{chen2024implicit}
Zhuangzhuang Chen, Ronghao Lu, Jie Chen, Houbing~Herbert Song, and Jianqiang Li.
\newblock Implicit gradient-modulated semantic data augmentation for deep crack recognition.
\newblock \emph{IEEE Transactions on Intelligent Transportation Systems}, 2024.

\bibitem[DeAndres-Tame et~al.(2024)DeAndres-Tame, Tolosana, Melzi, Vera-Rodriguez, Kim, Rathgeb, Liu, Morales, Fierrez, Ortega-Garcia, et~al.]{deandres2024frcsyn}
Ivan DeAndres-Tame, Ruben Tolosana, Pietro Melzi, Ruben Vera-Rodriguez, Minchul Kim, Christian Rathgeb, Xiaoming Liu, Aythami Morales, Julian Fierrez, Javier Ortega-Garcia, et~al.
\newblock Frcsyn challenge at cvpr 2024: Face recognition challenge in the era of synthetic data.
\newblock In \emph{Proceedings of the IEEE/CVF Conference on Computer Vision and Pattern Recognition}, pages 3173--3183, 2024.

\bibitem[DeAndres-Tame et~al.(2025)DeAndres-Tame, Tolosana, Melzi, Vera-Rodriguez, Kim, Rathgeb, Liu, Gomez, Morales, Fierrez, Ortega-Garcia, Zhong, Huang, Mi, Ding, Zhou, He, Fu, Cong, Zhang, Xiao, Smirnov, Pimenov, Grigorev, Timoshenko, Asfaw, Low, Liu, Wang, Zuo, He, Shahreza, George, Unnervik, Rahimi, Marcel, Neto, Huber, Kolf, Damer, Boutros, Cardoso, Sequeira, Atzori, Fenu, Marras, Štruc, Yu, Li, Li, Zhao, Lei, Zhu, Zhang, Biesseck, Vidal, Coelho, Granada, and Menotti]{frcsyn2025}
Ivan DeAndres-Tame, Ruben Tolosana, Pietro Melzi, Ruben Vera-Rodriguez, Minchul Kim, Christian Rathgeb, Xiaoming Liu, Luis~F. Gomez, Aythami Morales, Julian Fierrez, Javier Ortega-Garcia, Zhizhou Zhong, Yuge Huang, Yuxi Mi, Shouhong Ding, Shuigeng Zhou, Shuai He, Lingzhi Fu, Heng Cong, Rongyu Zhang, Zhihong Xiao, Evgeny Smirnov, Anton Pimenov, Aleksei Grigorev, Denis Timoshenko, Kaleb~Mesfin Asfaw, Cheng~Yaw Low, Hao Liu, Chuyi Wang, Qing Zuo, Zhixiang He, Hatef~Otroshi Shahreza, Anjith George, Alexander Unnervik, Parsa Rahimi, Sébastien Marcel, Pedro~C. Neto, Marco Huber, Jan~Niklas Kolf, Naser Damer, Fadi Boutros, Jaime~S. Cardoso, Ana~F. Sequeira, Andrea Atzori, Gianni Fenu, Mirko Marras, Vitomir Štruc, Jiang Yu, Zhangjie Li, Jichun Li, Weisong Zhao, Zhen Lei, Xiangyu Zhu, Xiao-Yu Zhang, Bernardo Biesseck, Pedro Vidal, Luiz Coelho, Roger Granada, and David Menotti.
\newblock Second frcsyn-ongoing: Winning solutions and post-challenge analysis to improve face recognition with synthetic data.
\newblock \emph{Information Fusion}, 120:\penalty0 103099, 2025.

\bibitem[Deng et~al.(2009)Deng, Dong, Socher, Li, Li, and Fei-Fei]{deng2009imagenet}
Jia Deng, Wei Dong, Richard Socher, Li-Jia Li, Kai Li, and Li Fei-Fei.
\newblock Imagenet: A large-scale hierarchical image database.
\newblock In \emph{2009 IEEE conference on computer vision and pattern recognition}, pages 248--255. Ieee, 2009.

\bibitem[Deng et~al.(2018)Deng, Cheng, Xue, Zhou, and Zafeiriou]{deng2018uv}
Jiankang Deng, Shiyang Cheng, Niannan Xue, Yuxiang Zhou, and Stefanos Zafeiriou.
\newblock Uv-gan: Adversarial facial uv map completion for pose-invariant face recognition.
\newblock In \emph{Proceedings of the IEEE conference on computer vision and pattern recognition}, pages 7093--7102, 2018.

\bibitem[Deng et~al.(2019)Deng, Guo, Xue, and Zafeiriou]{deng2019arcface}
Jiankang Deng, Jia Guo, Niannan Xue, and Stefanos Zafeiriou.
\newblock Arcface: Additive angular margin loss for deep face recognition.
\newblock In \emph{Proceedings of the IEEE/CVF conference on computer vision and pattern recognition}, pages 4690--4699, 2019.

\bibitem[Deng et~al.(2020)Deng, Yang, Chen, Wen, and Tong]{deng2020disentangled}
Yu Deng, Jiaolong Yang, Dong Chen, Fang Wen, and Xin Tong.
\newblock Disentangled and controllable face image generation via 3d imitative-contrastive learning.
\newblock In \emph{Proceedings of the IEEE/CVF conference on computer vision and pattern recognition}, pages 5154--5163, 2020.

\bibitem[Ding et~al.(2023)Ding, Zhang, Xia, Jebe, Tu, and Zhang]{ding2023diffusionrig}
Zheng Ding, Xuaner Zhang, Zhihao Xia, Lars Jebe, Zhuowen Tu, and Xiuming Zhang.
\newblock Diffusionrig: Learning personalized priors for facial appearance editing.
\newblock In \emph{Proceedings of the IEEE/CVF Conference on Computer Vision and Pattern Recognition}, pages 12736--12746, 2023.

\bibitem[Duta et~al.(2021)Duta, Liu, Zhu, and Shao]{duta2021improved}
Ionut~Cosmin Duta, Li Liu, Fan Zhu, and Ling Shao.
\newblock Improved residual networks for image and video recognition.
\newblock In \emph{2020 25th International Conference on Pattern Recognition (ICPR)}, pages 9415--9422. IEEE, 2021.

\bibitem[Feng et~al.(2021)Feng, Feng, Black, and Bolkart]{feng2021learning}
Yao Feng, Haiwen Feng, Michael~J Black, and Timo Bolkart.
\newblock Learning an animatable detailed 3d face model from in-the-wild images.
\newblock \emph{ACM Transactions on Graphics (ToG)}, 40\penalty0 (4):\penalty0 1--13, 2021.

\bibitem[Gal et~al.(2022)Gal, Alaluf, Atzmon, Patashnik, Bermano, Chechik, and Cohen-Or]{gal2022image}
Rinon Gal, Yuval Alaluf, Yuval Atzmon, Or Patashnik, Amit~H Bermano, Gal Chechik, and Daniel Cohen-Or.
\newblock An image is worth one word: Personalizing text-to-image generation using textual inversion.
\newblock \emph{arXiv preprint arXiv:2208.01618}, 2022.

\bibitem[Gal et~al.(2023)Gal, Arar, Atzmon, Bermano, Chechik, and Cohen-Or]{gal2023encoder}
Rinon Gal, Moab Arar, Yuval Atzmon, Amit~H Bermano, Gal Chechik, and Daniel Cohen-Or.
\newblock Encoder-based domain tuning for fast personalization of text-to-image models.
\newblock \emph{ACM Transactions on Graphics (TOG)}, 42\penalty0 (4):\penalty0 1--13, 2023.

\bibitem[Gecer et~al.(2018)Gecer, Bhattarai, Kittler, and Kim]{gecer2018semi}
Baris Gecer, Binod Bhattarai, Josef Kittler, and Tae-Kyun Kim.
\newblock Semi-supervised adversarial learning to generate photorealistic face images of new identities from 3d morphable model.
\newblock In \emph{Proceedings of the European conference on computer vision (ECCV)}, pages 217--234, 2018.

\bibitem[Geng et~al.(2019)Geng, Cao, and Tulyakov]{geng20193d}
Zhenglin Geng, Chen Cao, and Sergey Tulyakov.
\newblock 3d guided fine-grained face manipulation.
\newblock In \emph{Proceedings of the IEEE/CVF conference on computer vision and pattern recognition}, pages 9821--9830, 2019.

\bibitem[Guo et~al.(2016)Guo, Zhang, Hu, He, and Gao]{guo2016ms}
Yandong Guo, Lei Zhang, Yuxiao Hu, Xiaodong He, and Jianfeng Gao.
\newblock Ms-celeb-1m: A dataset and benchmark for large-scale face recognition.
\newblock In \emph{Computer Vision--ECCV 2016: 14th European Conference, Amsterdam, The Netherlands, October 11-14, 2016, Proceedings, Part III 14}, pages 87--102. Springer, 2016.

\bibitem[He et~al.(2016)He, Zhang, Ren, and Sun]{he2016deep}
Kaiming He, Xiangyu Zhang, Shaoqing Ren, and Jian Sun.
\newblock Deep residual learning for image recognition.
\newblock In \emph{Proceedings of the IEEE conference on computer vision and pattern recognition}, pages 770--778, 2016.

\bibitem[Ho and Salimans(2022)]{ho2022classifier}
Jonathan Ho and Tim Salimans.
\newblock Classifier-free diffusion guidance.
\newblock \emph{arXiv preprint arXiv:2207.12598}, 2022.

\bibitem[Ho et~al.(2020)Ho, Jain, and Abbeel]{ho2020denoising}
Jonathan Ho, Ajay Jain, and Pieter Abbeel.
\newblock Denoising diffusion probabilistic models.
\newblock \emph{Advances in neural information processing systems}, 33:\penalty0 6840--6851, 2020.

\bibitem[Hu et~al.(2021)Hu, Shen, Wallis, Allen-Zhu, Li, Wang, Wang, and Chen]{hu2021lora}
Edward~J Hu, Yelong Shen, Phillip Wallis, Zeyuan Allen-Zhu, Yuanzhi Li, Shean Wang, Lu Wang, and Weizhu Chen.
\newblock Lora: Low-rank adaptation of large language models.
\newblock \emph{arXiv preprint arXiv:2106.09685}, 2021.

\bibitem[Huang et~al.(2017)Huang, Liu, Van Der~Maaten, and Weinberger]{huang2017densely}
Gao Huang, Zhuang Liu, Laurens Van Der~Maaten, and Kilian~Q Weinberger.
\newblock Densely connected convolutional networks.
\newblock In \emph{Proceedings of the IEEE conference on computer vision and pattern recognition}, pages 4700--4708, 2017.

\bibitem[Huang et~al.(2008)Huang, Mattar, Berg, and Learned-Miller]{huang2008labeled}
Gary~B Huang, Marwan Mattar, Tamara Berg, and Eric Learned-Miller.
\newblock Labeled faces in the wild: A database forstudying face recognition in unconstrained environments.
\newblock In \emph{Workshop on faces in'Real-Life'Images: detection, alignment, and recognition}, 2008.

\bibitem[Huang et~al.(2023)Huang, Lang, Guo, He, Xue, Zhao, and Zhou]{huang2023dr}
Xianliang Huang, Yining Lang, Ying Guo, Yuan He, Hui Xue, Li Zhao, and Shuigeng Zhou.
\newblock Dr-net: A multi-view face synthesis network driven by dual representation.
\newblock In \emph{2023 IEEE International Conference on Multimedia and Expo (ICME)}, pages 1751--1756. IEEE, 2023.

\bibitem[Huang et~al.(2020)Huang, Wang, Tai, Liu, Shen, Li, Li, and Huang]{huang2020curricularface}
Yuge Huang, Yuhan Wang, Ying Tai, Xiaoming Liu, Pengcheng Shen, Shaoxin Li, Jilin Li, and Feiyue Huang.
\newblock Curricularface: adaptive curriculum learning loss for deep face recognition.
\newblock In \emph{proceedings of the IEEE/CVF conference on computer vision and pattern recognition}, pages 5901--5910, 2020.

\bibitem[Karras et~al.(2019)Karras, Laine, and Aila]{karras2019style}
Tero Karras, Samuli Laine, and Timo Aila.
\newblock A style-based generator architecture for generative adversarial networks.
\newblock In \emph{Proceedings of the IEEE/CVF conference on computer vision and pattern recognition}, pages 4401--4410, 2019.

\bibitem[Karras et~al.(2020{\natexlab{a}})Karras, Aittala, Hellsten, Laine, Lehtinen, and Aila]{karras2020training}
Tero Karras, Miika Aittala, Janne Hellsten, Samuli Laine, Jaakko Lehtinen, and Timo Aila.
\newblock Training generative adversarial networks with limited data.
\newblock \emph{Advances in neural information processing systems}, 33:\penalty0 12104--12114, 2020{\natexlab{a}}.

\bibitem[Karras et~al.(2020{\natexlab{b}})Karras, Laine, Aittala, Hellsten, Lehtinen, and Aila]{karras2020analyzing}
Tero Karras, Samuli Laine, Miika Aittala, Janne Hellsten, Jaakko Lehtinen, and Timo Aila.
\newblock Analyzing and improving the image quality of stylegan.
\newblock In \emph{Proceedings of the IEEE/CVF conference on computer vision and pattern recognition}, pages 8110--8119, 2020{\natexlab{b}}.

\bibitem[Karras et~al.(2021)Karras, Aittala, Laine, H{\"a}rk{\"o}nen, Hellsten, Lehtinen, and Aila]{karras2021alias}
Tero Karras, Miika Aittala, Samuli Laine, Erik H{\"a}rk{\"o}nen, Janne Hellsten, Jaakko Lehtinen, and Timo Aila.
\newblock Alias-free generative adversarial networks.
\newblock \emph{Advances in neural information processing systems}, 34:\penalty0 852--863, 2021.

\bibitem[Kemelmacher-Shlizerman et~al.(2016)Kemelmacher-Shlizerman, Seitz, Miller, and Brossard]{kemelmacher2016megaface}
Ira Kemelmacher-Shlizerman, Steven~M Seitz, Daniel Miller, and Evan Brossard.
\newblock The megaface benchmark: 1 million faces for recognition at scale.
\newblock In \emph{Proceedings of the IEEE conference on computer vision and pattern recognition}, pages 4873--4882, 2016.

\bibitem[Kim et~al.(2018)Kim, Garrido, Tewari, Xu, Thies, Niessner, P{\'e}rez, Richardt, Zollh{\"o}fer, and Theobalt]{kim2018deep}
Hyeongwoo Kim, Pablo Garrido, Ayush Tewari, Weipeng Xu, Justus Thies, Matthias Niessner, Patrick P{\'e}rez, Christian Richardt, Michael Zollh{\"o}fer, and Christian Theobalt.
\newblock Deep video portraits.
\newblock \emph{ACM transactions on graphics (TOG)}, 37\penalty0 (4):\penalty0 1--14, 2018.

\bibitem[Kim et~al.(2022)Kim, Jain, and Liu]{kim2022adaface}
Minchul Kim, Anil~K. Jain, and Xiaoming Liu.
\newblock Adaface: Quality adaptive margin for face recognition.
\newblock In \emph{Proceedings of the IEEE/CVF Conference on Computer Vision and Pattern Recognition (CVPR)}, pages 18750--18759, 2022.

\bibitem[Kim et~al.(2023)Kim, Liu, Jain, and Liu]{kim2023dcface}
Minchul Kim, Feng Liu, Anil Jain, and Xiaoming Liu.
\newblock Dcface: Synthetic face generation with dual condition diffusion model.
\newblock In \emph{Proceedings of the ieee/cvf conference on computer vision and pattern recognition}, pages 12715--12725, 2023.

\bibitem[Kingma(2014)]{kingma2014adam}
Diederik~P Kingma.
\newblock Adam: A method for stochastic optimization.
\newblock \emph{arXiv preprint arXiv:1412.6980}, 2014.

\bibitem[Kolf et~al.(2023)Kolf, Rieber, Elliesen, Boutros, Kuijper, and Damer]{kolf2023identity}
Jan~Niklas Kolf, Tim Rieber, Jurek Elliesen, Fadi Boutros, Arjan Kuijper, and Naser Damer.
\newblock Identity-driven three-player generative adversarial network for synthetic-based face recognition.
\newblock In \emph{Proceedings of the IEEE/CVF Conference on Computer Vision and Pattern Recognition}, pages 806--816, 2023.

\bibitem[Kynk{\"a}{\"a}nniemi et~al.(2019)Kynk{\"a}{\"a}nniemi, Karras, Laine, Lehtinen, and Aila]{kynkaanniemi2019improved}
Tuomas Kynk{\"a}{\"a}nniemi, Tero Karras, Samuli Laine, Jaakko Lehtinen, and Timo Aila.
\newblock Improved precision and recall metric for assessing generative models.
\newblock \emph{Advances in neural information processing systems}, 32, 2019.

\bibitem[Learned-Miller(2014)]{lfwtechupdate}
Gary B. Huang~Erik Learned-Miller.
\newblock Labeled faces in the wild: Updates and new reporting procedures.
\newblock Technical Report UM-CS-2014-003, University of Massachusetts, Amherst, 2014.

\bibitem[Li et~al.(2021)Li, Chen, Chen, and Lin]{8972606}
Jianqiang Li, Zhuangzhuang Chen, Jie Chen, and Qiuzhen Lin.
\newblock Diversity-sensitive generative adversarial network for terrain mapping under limited human intervention.
\newblock \emph{IEEE Transactions on Cybernetics}, 51\penalty0 (12):\penalty0 6029--6040, 2021.

\bibitem[Li et~al.(2024{\natexlab{a}})Li, Xu, Wu, Xiong, Deng, Ji, Huang, Feng, Ding, and Hooi]{li2024id}
Shen Li, Jianqing Xu, Jiaying Wu, Miao Xiong, Ailin Deng, Jiazhen Ji, Yuge Huang, Wenjie Feng, Shouhong Ding, and Bryan Hooi.
\newblock Id3: Identity-preserving-yet-diversified diffusion models for synthetic face recognition.
\newblock \emph{arXiv preprint arXiv:2409.17576}, 2024{\natexlab{a}}.

\bibitem[Li et~al.(2017)Li, Bolkart, Black, Li, and Romero]{li2017learning}
Tianye Li, Timo Bolkart, Michael~J Black, Hao Li, and Javier Romero.
\newblock Learning a model of facial shape and expression from 4d scans.
\newblock \emph{ACM Trans. Graph.}, 36\penalty0 (6):\penalty0 194--1, 2017.

\bibitem[Li et~al.(2024{\natexlab{b}})Li, Zhang, Kang, Cheng, Gao, Zhang, Liang, Liao, Cao, and Shan]{li2024advances}
Xiaoyu Li, Qi Zhang, Di Kang, Weihao Cheng, Yiming Gao, Jingbo Zhang, Zhihao Liang, Jing Liao, Yan-Pei Cao, and Ying Shan.
\newblock Advances in 3d generation: A survey.
\newblock \emph{arXiv preprint arXiv:2401.17807}, 2024{\natexlab{b}}.

\bibitem[Li et~al.(2024{\natexlab{c}})Li, Cao, Wang, Qi, Cheng, and Shan]{li2024photomaker}
Zhen Li, Mingdeng Cao, Xintao Wang, Zhongang Qi, Ming-Ming Cheng, and Ying Shan.
\newblock Photomaker: Customizing realistic human photos via stacked id embedding.
\newblock In \emph{Proceedings of the IEEE/CVF Conference on Computer Vision and Pattern Recognition}, pages 8640--8650, 2024{\natexlab{c}}.

\bibitem[Medin et~al.(2022)Medin, Egger, Cherian, Wang, Tenenbaum, Liu, and Marks]{medin2022most}
Safa~C Medin, Bernhard Egger, Anoop Cherian, Ye Wang, Joshua~B Tenenbaum, Xiaoming Liu, and Tim~K Marks.
\newblock Most-gan: 3d morphable stylegan for disentangled face image manipulation.
\newblock In \emph{Proceedings of the AAAI conference on artificial intelligence}, pages 1962--1971, 2022.

\bibitem[Melzi et~al.(2024)Melzi, Tolosana, Vera-Rodriguez, Kim, Rathgeb, Liu, DeAndres-Tame, Morales, Fierrez, Ortega-Garcia, et~al.]{melzi2024frcsyn}
Pietro Melzi, Ruben Tolosana, Ruben Vera-Rodriguez, Minchul Kim, Christian Rathgeb, Xiaoming Liu, Ivan DeAndres-Tame, Aythami Morales, Julian Fierrez, Javier Ortega-Garcia, et~al.
\newblock Frcsyn challenge at wacv 2024: Face recognition challenge in the era of synthetic data.
\newblock In \emph{Proceedings of the IEEE/CVF Winter Conference on Applications of Computer Vision}, pages 892--901, 2024.

\bibitem[Mi et~al.(2022)Mi, Huang, Ji, Liu, Xu, Ding, and Zhou]{mi2022duetface}
Yuxi Mi, Yuge Huang, Jiazhen Ji, Hongquan Liu, Xingkun Xu, Shouhong Ding, and Shuigeng Zhou.
\newblock Duetface: Collaborative privacy-preserving face recognition via channel splitting in the frequency domain.
\newblock In \emph{Proceedings of the 30th ACM International Conference on Multimedia}, pages 6755--6764, 2022.

\bibitem[Mi et~al.(2024)Mi, Zhong, Huang, Ji, Xu, Wang, Wang, Ding, and Zhou]{mi2024privacy}
Yuxi Mi, Zhizhou Zhong, Yuge Huang, Jiazhen Ji, Jianqing Xu, Jun Wang, Shaoming Wang, Shouhong Ding, and Shuigeng Zhou.
\newblock Privacy-preserving face recognition using trainable feature subtraction.
\newblock In \emph{Proceedings of the IEEE/CVF Conference on Computer Vision and Pattern Recognition}, pages 297--307, 2024.

\bibitem[Moschoglou et~al.(2017)Moschoglou, Papaioannou, Sagonas, Deng, Kotsia, and Zafeiriou]{moschoglou2017agedb}
Stylianos Moschoglou, Athanasios Papaioannou, Christos Sagonas, Jiankang Deng, Irene Kotsia, and Stefanos Zafeiriou.
\newblock Agedb: the first manually collected, in-the-wild age database.
\newblock In \emph{proceedings of the IEEE conference on computer vision and pattern recognition workshops}, pages 51--59, 2017.

\bibitem[Nguyen-Phuoc et~al.(2019)Nguyen-Phuoc, Li, Theis, Richardt, and Yang]{nguyen2019hologan}
Thu Nguyen-Phuoc, Chuan Li, Lucas Theis, Christian Richardt, and Yong-Liang Yang.
\newblock Hologan: Unsupervised learning of 3d representations from natural images.
\newblock In \emph{Proceedings of the IEEE/CVF International Conference on Computer Vision}, pages 7588--7597, 2019.

\bibitem[Ou et~al.(2021)Ou, Chen, Zhang, Huang, Li, Li, Li, Cao, and Wang]{ou2021sdd}
Fu-Zhao Ou, Xingyu Chen, Ruixin Zhang, Yuge Huang, Shaoxin Li, Jilin Li, Yong Li, Liujuan Cao, and Yuan-Gen Wang.
\newblock Sdd-fiqa: Unsupervised face image quality assessment with similarity distribution distance.
\newblock In \emph{Proceedings of the IEEE/CVF conference on computer vision and pattern recognition}, pages 7670--7679, 2021.

\bibitem[Papantoniou et~al.(2024)Papantoniou, Lattas, Moschoglou, Deng, Kainz, and Zafeiriou]{papantoniou2024arc2face}
Foivos~Paraperas Papantoniou, Alexandros Lattas, Stylianos Moschoglou, Jiankang Deng, Bernhard Kainz, and Stefanos Zafeiriou.
\newblock Arc2face: A foundation model of human faces.
\newblock \emph{arXiv preprint arXiv:2403.11641}, 2024.

\bibitem[Paysan et~al.(2009)Paysan, Knothe, Amberg, Romdhani, and Vetter]{paysan20093d}
Pascal Paysan, Reinhard Knothe, Brian Amberg, Sami Romdhani, and Thomas Vetter.
\newblock A 3d face model for pose and illumination invariant face recognition.
\newblock In \emph{2009 sixth IEEE international conference on advanced video and signal based surveillance}, pages 296--301. Ieee, 2009.

\bibitem[Peng et~al.(2024)Peng, Zhu, Jiang, Tai, Luo, Zhang, Lin, Jin, Wang, and Ji]{peng2024portraitbooth}
Xu Peng, Junwei Zhu, Boyuan Jiang, Ying Tai, Donghao Luo, Jiangning Zhang, Wei Lin, Taisong Jin, Chengjie Wang, and Rongrong Ji.
\newblock Portraitbooth: A versatile portrait model for fast identity-preserved personalization.
\newblock In \emph{Proceedings of the IEEE/CVF Conference on Computer Vision and Pattern Recognition}, pages 27080--27090, 2024.

\bibitem[Piao et~al.(2019)Piao, Qian, and Li]{piao2019semi}
Jingtan Piao, Chen Qian, and Hongsheng Li.
\newblock Semi-supervised monocular 3d face reconstruction with end-to-end shape-preserved domain transfer.
\newblock In \emph{Proceedings of the IEEE/CVF international conference on computer vision}, pages 9398--9407, 2019.

\bibitem[Qian et~al.(2024)Qian, Cai, Pan, Li, Yao, Sun, and Mei]{qian2024boosting}
Yurui Qian, Qi Cai, Yingwei Pan, Yehao Li, Ting Yao, Qibin Sun, and Tao Mei.
\newblock Boosting diffusion models with moving average sampling in frequency domain.
\newblock In \emph{Proceedings of the IEEE/CVF Conference on Computer Vision and Pattern Recognition}, pages 8911--8920, 2024.

\bibitem[Qiu et~al.(2021)Qiu, Yu, Gong, Li, Liu, and Tao]{qiu2021synface}
Haibo Qiu, Baosheng Yu, Dihong Gong, Zhifeng Li, Wei Liu, and Dacheng Tao.
\newblock Synface: Face recognition with synthetic data.
\newblock In \emph{Proceedings of the IEEE/CVF International Conference on Computer Vision}, pages 10880--10890, 2021.

\bibitem[Radford et~al.(2021)Radford, Kim, Hallacy, Ramesh, Goh, Agarwal, Sastry, Askell, Mishkin, Clark, et~al.]{radford2021learning}
Alec Radford, Jong~Wook Kim, Chris Hallacy, Aditya Ramesh, Gabriel Goh, Sandhini Agarwal, Girish Sastry, Amanda Askell, Pamela Mishkin, Jack Clark, et~al.
\newblock Learning transferable visual models from natural language supervision.
\newblock In \emph{International conference on machine learning}, pages 8748--8763. PMLR, 2021.

\bibitem[Rombach et~al.(2022)Rombach, Blattmann, Lorenz, Esser, and Ommer]{rombach2022high}
Robin Rombach, Andreas Blattmann, Dominik Lorenz, Patrick Esser, and Bj{\"o}rn Ommer.
\newblock High-resolution image synthesis with latent diffusion models.
\newblock In \emph{Proceedings of the IEEE/CVF conference on computer vision and pattern recognition}, pages 10684--10695, 2022.

\bibitem[Ronneberger et~al.(2015)Ronneberger, Fischer, and Brox]{ronneberger2015u}
Olaf Ronneberger, Philipp Fischer, and Thomas Brox.
\newblock U-net: Convolutional networks for biomedical image segmentation.
\newblock In \emph{Medical image computing and computer-assisted intervention--MICCAI 2015: 18th international conference, Munich, Germany, October 5-9, 2015, proceedings, part III 18}, pages 234--241. Springer, 2015.

\bibitem[Ruder(2016)]{ruder2016overview}
Sebastian Ruder.
\newblock An overview of gradient descent optimization algorithms.
\newblock \emph{arXiv preprint arXiv:1609.04747}, 2016.

\bibitem[Ruhe et~al.(2024)Ruhe, Heek, Salimans, and Hoogeboom]{ruhe2024rolling}
David Ruhe, Jonathan Heek, Tim Salimans, and Emiel Hoogeboom.
\newblock Rolling diffusion models.
\newblock \emph{arXiv preprint arXiv:2402.09470}, 2024.

\bibitem[Ruiz et~al.(2023)Ruiz, Li, Jampani, Pritch, Rubinstein, and Aberman]{ruiz2023dreambooth}
Nataniel Ruiz, Yuanzhen Li, Varun Jampani, Yael Pritch, Michael Rubinstein, and Kfir Aberman.
\newblock Dreambooth: Fine tuning text-to-image diffusion models for subject-driven generation.
\newblock In \emph{Proceedings of the IEEE/CVF conference on computer vision and pattern recognition}, pages 22500--22510, 2023.

\bibitem[Salimans et~al.(2016)Salimans, Goodfellow, Zaremba, Cheung, Radford, and Chen]{salimans2016improved}
Tim Salimans, Ian Goodfellow, Wojciech Zaremba, Vicki Cheung, Alec Radford, and Xi Chen.
\newblock Improved techniques for training gans.
\newblock \emph{Advances in neural information processing systems}, 29, 2016.

\bibitem[Sengupta et~al.(2016)Sengupta, Chen, Castillo, Patel, Chellappa, and Jacobs]{sengupta2016frontal}
Soumyadip Sengupta, Jun-Cheng Chen, Carlos Castillo, Vishal~M Patel, Rama Chellappa, and David~W Jacobs.
\newblock Frontal to profile face verification in the wild.
\newblock In \emph{2016 IEEE winter conference on applications of computer vision (WACV)}, pages 1--9. IEEE, 2016.

\bibitem[Shen et~al.(2018{\natexlab{a}})Shen, Luo, Yan, Wang, and Tang]{shen2018faceid}
Yujun Shen, Ping Luo, Junjie Yan, Xiaogang Wang, and Xiaoou Tang.
\newblock Faceid-gan: Learning a symmetry three-player gan for identity-preserving face synthesis.
\newblock In \emph{Proceedings of the IEEE conference on computer vision and pattern recognition}, pages 821--830, 2018{\natexlab{a}}.

\bibitem[Shen et~al.(2018{\natexlab{b}})Shen, Zhou, Luo, and Tang]{shen2018facefeat}
Yujun Shen, Bolei Zhou, Ping Luo, and Xiaoou Tang.
\newblock Facefeat-gan: a two-stage approach for identity-preserving face synthesis.
\newblock \emph{arXiv preprint arXiv:1812.01288}, 2018{\natexlab{b}}.

\bibitem[Shen et~al.(2020)Shen, Gu, Tang, and Zhou]{shen2020interpreting}
Yujun Shen, Jinjin Gu, Xiaoou Tang, and Bolei Zhou.
\newblock Interpreting the latent space of gans for semantic face editing.
\newblock In \emph{Proceedings of the IEEE/CVF conference on computer vision and pattern recognition}, pages 9243--9252, 2020.

\bibitem[Song et~al.(2020)Song, Meng, and Ermon]{song2020denoising}
Jiaming Song, Chenlin Meng, and Stefano Ermon.
\newblock Denoising diffusion implicit models.
\newblock \emph{arXiv preprint arXiv:2010.02502}, 2020.

\bibitem[Sun et~al.(2024)Sun, Song, Patras, and Tzimiropoulos]{sun2024cemiface}
Zhonglin Sun, Siyang Song, Ioannis Patras, and Georgios Tzimiropoulos.
\newblock Cemiface: Center-based semi-hard synthetic face generation for face recognition.
\newblock \emph{arXiv preprint arXiv:2409.18876}, 2024.

\bibitem[Tran et~al.(2018)Tran, Yin, and Liu]{tran2018representation}
Luan Tran, Xi Yin, and Xiaoming Liu.
\newblock Representation learning by rotating your faces.
\newblock \emph{IEEE transactions on pattern analysis and machine intelligence}, 41\penalty0 (12):\penalty0 3007--3021, 2018.

\bibitem[Valevski et~al.(2023)Valevski, Lumen, Matias, and Leviathan]{valevski2023face0}
Dani Valevski, Danny Lumen, Yossi Matias, and Yaniv Leviathan.
\newblock Face0: Instantaneously conditioning a text-to-image model on a face.
\newblock In \emph{SIGGRAPH Asia 2023 Conference Papers}, pages 1--10, 2023.

\bibitem[Wang et~al.(2018)Wang, Wang, Zhou, Ji, Gong, Zhou, Li, and Liu]{wang2018cosface}
Hao Wang, Yitong Wang, Zheng Zhou, Xing Ji, Dihong Gong, Jingchao Zhou, Zhifeng Li, and Wei Liu.
\newblock Cosface: Large margin cosine loss for deep face recognition.
\newblock In \emph{Proceedings of the IEEE conference on computer vision and pattern recognition}, pages 5265--5274, 2018.

\bibitem[Wang et~al.(2024)Wang, Jia, Li, Li, Ma, Zhuge, and Lu]{wang2024stableidentity}
Qinghe Wang, Xu Jia, Xiaomin Li, Taiqing Li, Liqian Ma, Yunzhi Zhuge, and Huchuan Lu.
\newblock Stableidentity: Inserting anybody into anywhere at first sight.
\newblock \emph{arXiv preprint arXiv:2401.15975}, 2024.

\bibitem[Wang et~al.(2022)Wang, Liu, Luo, Yang, and Wang]{wang2022privacy}
Yinggui Wang, Jian Liu, Man Luo, Le Yang, and Li Wang.
\newblock Privacy-preserving face recognition in the frequency domain.
\newblock In \emph{Proceedings of the AAAI Conference on Artificial Intelligence}, pages 2558--2566, 2022.

\bibitem[Xiao et~al.(2024)Xiao, Yin, Freeman, Durand, and Han]{xiao2024fastcomposer}
Guangxuan Xiao, Tianwei Yin, William~T Freeman, Fr{\'e}do Durand, and Song Han.
\newblock Fastcomposer: Tuning-free multi-subject image generation with localized attention.
\newblock \emph{International Journal of Computer Vision}, pages 1--20, 2024.

\bibitem[Xu et~al.(2024)Xu, Zhang, Yang, Li, and Li]{xu2024chain}
Zunnan Xu, Yachao Zhang, Sicheng Yang, Ronghui Li, and Xiu Li.
\newblock Chain of generation: Multi-modal gesture synthesis via cascaded conditional control.
\newblock In \emph{Proceedings of the AAAI Conference on Artificial Intelligence}, pages 6387--6395, 2024.

\bibitem[Yan et~al.(2023)Yan, Zhang, Wang, Zhou, Zhang, Cheng, Yu, and Fu]{yan2023facestudio}
Yuxuan Yan, Chi Zhang, Rui Wang, Yichao Zhou, Gege Zhang, Pei Cheng, Gang Yu, and Bin Fu.
\newblock Facestudio: Put your face everywhere in seconds.
\newblock \emph{arXiv preprint arXiv:2312.02663}, 2023.

\bibitem[Yan et~al.(2024{\natexlab{a}})Yan, Wang, Wang, Jin, Zhang, Chen, Yao, Ding, Wu, and Yuan]{yan2024effort}
Zhiyuan Yan, Jiangming Wang, Zhendong Wang, Peng Jin, Ke-Yue Zhang, Shen Chen, Taiping Yao, Shouhong Ding, Baoyuan Wu, and Li Yuan.
\newblock Effort: Efficient orthogonal modeling for generalizable ai-generated image detection.
\newblock \emph{arXiv preprint arXiv:2411.15633}, 2024{\natexlab{a}}.

\bibitem[Yan et~al.(2024{\natexlab{b}})Yan, Yao, Chen, Zhao, Fu, Zhu, Luo, Wang, Ding, Wu, et~al.]{yan2024df40}
Zhiyuan Yan, Taiping Yao, Shen Chen, Yandan Zhao, Xinghe Fu, Junwei Zhu, Donghao Luo, Chengjie Wang, Shouhong Ding, Yunsheng Wu, et~al.
\newblock Df40: Toward next-generation deepfake detection.
\newblock \emph{arXiv preprint arXiv:2406.13495}, 2024{\natexlab{b}}.

\bibitem[Yi et~al.(2014)Yi, Lei, Liao, and Li]{yi2014learning}
Dong Yi, Zhen Lei, Shengcai Liao, and Stan~Z Li.
\newblock Learning face representation from scratch.
\newblock \emph{arXiv preprint arXiv:1411.7923}, 2014.

\bibitem[Yuan et~al.(2023)Yuan, Cun, Zhang, Li, Qi, Wang, Shan, and Zheng]{yuan2023inserting}
Ge Yuan, Xiaodong Cun, Yong Zhang, Maomao Li, Chenyang Qi, Xintao Wang, Ying Shan, and Huicheng Zheng.
\newblock Inserting anybody in diffusion models via celeb basis.
\newblock \emph{arXiv preprint arXiv:2306.00926}, 2023.

\bibitem[Zhang et~al.(2018)Zhang, Isola, Efros, Shechtman, and Wang]{zhang2018unreasonable}
Richard Zhang, Phillip Isola, Alexei~A Efros, Eli Shechtman, and Oliver Wang.
\newblock The unreasonable effectiveness of deep features as a perceptual metric.
\newblock In \emph{Proceedings of the IEEE conference on computer vision and pattern recognition}, pages 586--595, 2018.

\bibitem[Zheng and Deng(2018)]{zheng2018cross}
Tianyue Zheng and Weihong Deng.
\newblock Cross-pose lfw: A database for studying cross-pose face recognition in unconstrained environments.
\newblock \emph{Beijing University of Posts and Telecommunications, Tech. Rep}, 5\penalty0 (7):\penalty0 5, 2018.

\bibitem[Zheng et~al.(2017)Zheng, Deng, and Hu]{zheng2017cross}
Tianyue Zheng, Weihong Deng, and Jiani Hu.
\newblock Cross-age lfw: A database for studying cross-age face recognition in unconstrained environments.
\newblock \emph{arXiv preprint arXiv:1708.08197}, 2017.

\bibitem[Zhong et~al.(2024)Zhong, Mi, Huang, Xu, Mu, Ding, Zhang, Guo, Wu, and Zhou]{zhong2024slerpface}
Zhizhou Zhong, Yuxi Mi, Yuge Huang, Jianqing Xu, Guodong Mu, Shouhong Ding, Jingyun Zhang, Rizen Guo, Yunsheng Wu, and Shuigeng Zhou.
\newblock Slerpface: face template protection via spherical linear interpolation.
\newblock \emph{arXiv preprint arXiv:2407.03043}, 2024.

\bibitem[Zhou et~al.(2023)Zhou, Zhang, Sun, and Xu]{zhou2023enhancing}
Yufan Zhou, Ruiyi Zhang, Tong Sun, and Jinhui Xu.
\newblock Enhancing detail preservation for customized text-to-image generation: A regularization-free approach.
\newblock \emph{arXiv preprint arXiv:2305.13579}, 2023.

\bibitem[Zhu et~al.(2021)Zhu, Huang, Deng, Ye, Huang, Chen, Zhu, Yang, Lu, Du, et~al.]{zhu2021webface260m}
Zheng Zhu, Guan Huang, Jiankang Deng, Yun Ye, Junjie Huang, Xinze Chen, Jiagang Zhu, Tian Yang, Jiwen Lu, Dalong Du, et~al.
\newblock Webface260m: A benchmark unveiling the power of million-scale deep face recognition.
\newblock In \emph{Proceedings of the IEEE/CVF Conference on Computer Vision and Pattern Recognition}, pages 10492--10502, 2021.

\end{thebibliography}
